\documentclass[journal]{IEEEtran}
\usepackage{graphicx}
\usepackage{cite}
\usepackage{picinpar}
\usepackage{amsmath,bm}
\usepackage{url}
\usepackage[latin1]{inputenc}
\usepackage{colortbl}
\usepackage{soul}
\usepackage{multirow}
\usepackage{pifont}
\usepackage{color}
\usepackage{alltt}
\usepackage[hidelinks]{hyperref}
\usepackage{enumerate}
\usepackage{siunitx}
\usepackage{breakurl}
\usepackage{epstopdf}
\usepackage{pbox}
\usepackage{indentfirst}
\usepackage{booktabs}
\usepackage{makecell}
\usepackage{subfigure}
\usepackage{float}
\usepackage{threeparttable}
\usepackage[numbers,sort&compress]{natbib}

\begin{document}

\title{Wheel-INS: A Wheel-mounted MEMS IMU-based Dead Reckoning System}

\author{Xiaoji~Niu, Yibin~Wu, and~Jian~Kuang% <-this % stops a space
\thanks{This work was funded in part by the National Key Research and Development Program of China (No. 2016YFB0501800 and No. 2016YFB0502202).}% <-this % stops a space
\thanks{The authors are with the GNSS Research Center, Wuhan University, Wuhan, China. \{xjniu, ybwu, kuang\}@whu.edu.cn.}}% <-this % stops a space

\maketitle

\begin{abstract}
To improve the accuracy and robustness of the inertial navigation systems (INS) for wheeled robots without adding additional component cost, we propose Wheel-INS, a complete dead reckoning solution based on a wheel-mounted microelectromechanical system (MEMS) inertial measurement unit (IMU). There are two major advantages by mounting an IMU to the center of a non-steering wheel of the ground vehicle. Firstly, the gyroscope outputs can be used to calculate the wheel speed, so as to replace the traditional odometer to mitigate the error drift of INS. Secondly, with the rotation of the wheel, the constant bias error of the inertial sensor can be canceled to some extent. The installation scheme of the wheel-mounted IMU (Wheel-IMU), the system characteristics, and the dead reckoning error analysis are described. Experimental results show that the maximum position drift of Wheel-INS in the horizontal plane is less than 1.8\% of the total traveled distance, reduced by 23\% compared to the conventional odometer-aided INS (ODO/INS). In addition, Wheel-INS outperforms ODO/INS because of its inherent immunity to constant bias error of gyroscopes. The source code and experimental datasets used in this paper is made available to the community (https://github.com/i2Nav-WHU/Wheel-INS).
\end{abstract}


\IEEEpeerreviewmaketitle

\section*{Nomenclature}
\begin{enumerate}[a)]
	\item Matrices are denoted as uppercase bold letters.
	\item Vectors are denoted as lowercase bold italic letters.
	\item Scalars are denoted as lowercase italic letters.
	\item Coordinate frames involved in the vector transformation are denoted as superscript and subscript. For vectors, the superscript denotes the projected coordinate system.
	\item $\hat{*}$, estimated or computed values.
	\item $\widetilde{*}$, observed or measured values.
	\item $\bm{a}_x$, element of vector $\bm{a}$ in the $x$ axis.
	
\end{enumerate}

\section{Introduction}
Positioning and heading estimation is essential for autonomous mobile robots as it provides important information for the navigation and control loop \cite{Kuutti2018}. Current position tracking systems have relied heavily on the Global Navigation Satellite System (GNSS) and other radio signal-based techniques \cite{grewal2020}. Although GNSS can provide centimeter-level \cite{teunissen2017springer} positioning results in open-sky scenarios, it deteriorates in complex environments such as urban canyons and forests owing to multipath and signal blockage. Moreover, GNSS is completely unavailable for indoor navigation. Therefore, investigation of relative positioning methods to bridge GNSS outages is crucial to improve the robustness and reliability of the navigation system of mobile robots \cite{zekavat2019}.

The inertial navigation system (INS) is a self-contained approach, exhibiting significant superiority when considering system tolerance to disturbance in the surrounding environment. With the rapid development of microelectromechanical system (MEMS) technology, MEMS inertial measurement units (IMU) have found wide application for motion detection, robot navigation, pedestrian dead reckoning and more because of their small size, low cost, light weight, and low power consumption \cite{du2015}. Nonetheless, because of the significant noise and bias instability of MEMS IMU, the positioning errors of INS drift quickly with time. Hence, aiding information is required to suppress the error accumulation of INS. 

The wheel odometer, a common sensor for ground wheeled robots, is extensively utilized to provide either distance or velocity information of the vehicle to suppress INS error drift thanks to its short-distance stability \cite{Shin2005, wu2010}. It has been proven that odometer and non-holonomic constraints (NHC) contribute significantly to restrain both the positioning and attitude errors and enhance the stability of INS \cite{godha2006, dissanayake2001}. However, reliability of the odometer data depends on road conditions and vehicle maneuvers, which degenerates if relative slippage occurs between the tires and contacting surface \cite{wu2010}. In addition, fusing information from different systems is challenging because of different standards, hardware modification, data transfer synchronization, and difficulties in obtaining reliability information along with the data \cite{collin2014tvt}. 

\textit{Can we use only one sensor modal to implement the similar information fusion scheme as the conventional odometer-aided INS (ODO/INS)?} The answer is \textit{yes}. By mounting the IMU to the wheel center, the wheel velocity can be calculated with the gyroscope outputs and the wheel radius, replacing the traditional wheel encoder or odometer. Furthermore, rotating the IMU around a certain axis with a fixed speed can modulate the constant bias of inertial sensors into sinusoidal signal which can be canceled after integral over one rotation period, i.e., rotary INS \cite{du2015, geller1968}. Traditionally, an additional heavy, expensive, and sophisticated controllable rotation platform is required to perform the rotary INS, which negates the advantages. Note that the wheels of a wheeled robot are inherent rotation devices. Although the rotation rate of the robot wheel cannot be controlled precisely, it can be approximately considered as constant during one cycle. Therefore, such a wheel-mounted MEMS IMU (Wheel-IMU) can be regarded as a noninvasive sensor which can take advantages of the rotation modulation and obtain the wheel velocity to provide high-accuracy self-contained state estimates without additional cost.

%As a result, improving the localization performance of IMU alone based dead reckoning system is highly desirable. It is worth mentioning that properly schemed motion of the IMU can effectively limit the error accumulation of INS, such as the carouseling IMU \cite{Groves2013} and rotary INS \cite{du2015, geller1968}. 

Few studies have been conducted with a focus on the Wheel-IMU based state estimators. The most representative solution is a 2-dimensional (2D) dead reckoning (DR) system developed in \cite{collin2014tvt, collin2014tim}. In this system, the data of the two accelerometers perpendicular to the rotation axis are used to determine the wheel rotation angle. Then, this angle is multiplied by the wheel radius to obtain the traveled distance. Gyroscope outputs are utilized to calculate the vehicle heading. Although this system has evolved into a commercial product, there are two major deficiencies. Firstly, the algorithm relies on the assumption of uniform vehicle motion, which is susceptible to continuous acceleration. Secondly, the misalignment errors between the IMU and the wheel center are not addressed.

\begin{figure}[htbp]
	\centering
	\includegraphics[width=8.8cm]{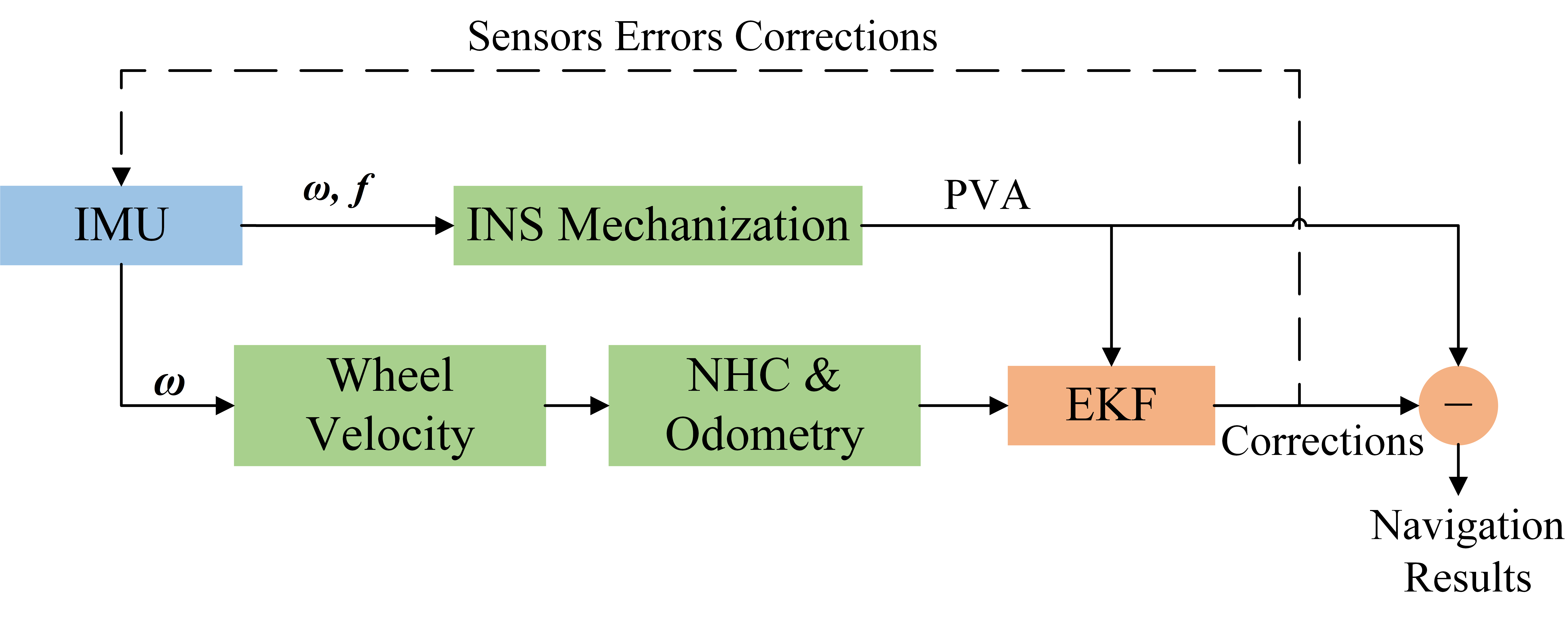}
	\caption{Overview of the structure of Wheel-INS. $\bm\omega$ and $\bm{f}$  are the angular rate and specific force measured by Wheel-IMU, respectively; PVA represents the position, velocity, and attitude of Wheel-IMU.}
	\label{fig1}
\end{figure}

In this study, a complete Wheel-IMU based DR system (Wheel-INS) is proposed. Fig. \ref{fig1} shows the system structure. Note that the DR system mentioned in this paper refers to the general method of calculating the relative vehicle position with the pose increment (e.g., INS), not just the traditional traveled distance and heading-based 2D DR system. In Wheel-INS, the forward INS mechanization is performed to predict the state of the vehicle. At the same time, the wheel velocity calculated by the gyroscope outputs and wheel radius is treated as an external observation with NHC to update the state. For the sake of simplicity and efficiency, we use the error-state extended Kalman filter (EKF) to implement the information fusion and state estimation. The state corrections estimated by the filter are fed back to update the vehicle pose and compensate the IMU outputs. Particularly, the main contributions of this article are:
\begin{itemize}
	\item A complete DR system based on a wheel-mounted IMU is proposed and implemented, where the gyroscope data are used to calculate the wheel speed to replace the conventional odometer.
	\item The installation scheme of Wheel-IMU is explained and the misalignment errors between Wheel-IMU and the wheel are defined and analyzed.
	\item We illustrate that the position and heading accuracy of Wheel-INS is much higher than ODO/INS through extensive field experiments. Furthermore, we show that Wheel-INS exhibits significant advantage in terms of immunity to constant gyroscope bias error.
\end{itemize}

The remainder of this paper is organized as follows. In Section \uppercase\expandafter{\romannumeral2}, the proposed installation scheme of Wheel-IMU and its rotation characteristics are described firstly; then, the misalignment errors between Wheel-IMU and the wheel are defined and analyzed. Implementation details of Wheel-INS, including the error state model and observation model, are presented in Section \uppercase\expandafter{\romannumeral3}. Experimental results are explained and discussed in Section \uppercase\expandafter{\romannumeral4}. Section \uppercase\expandafter{\romannumeral5} provides some conclusions and directions for future work.

\section{Prerequisites}
Unlike the conventional ODO/INS in which the IMU is placed on the vehicle body or in the trunk, in the proposed design, the IMU is mounted on the wheel to take advantages of the inherent rotation platform of the vehicle. In this section, the installation scheme of Wheel-IMU and the coordinate systems involved in Wheel-INS are introduced. Subsequently, the dynamic characteristics of Wheel-IMU and the misalignment errors between the wheel and Wheel-IMU are defined and analyzed. 
\subsection{Installation Scheme and Coordinate Systems}
\begin{figure}[htbp]
	\centering
	\includegraphics[width=8.8cm]{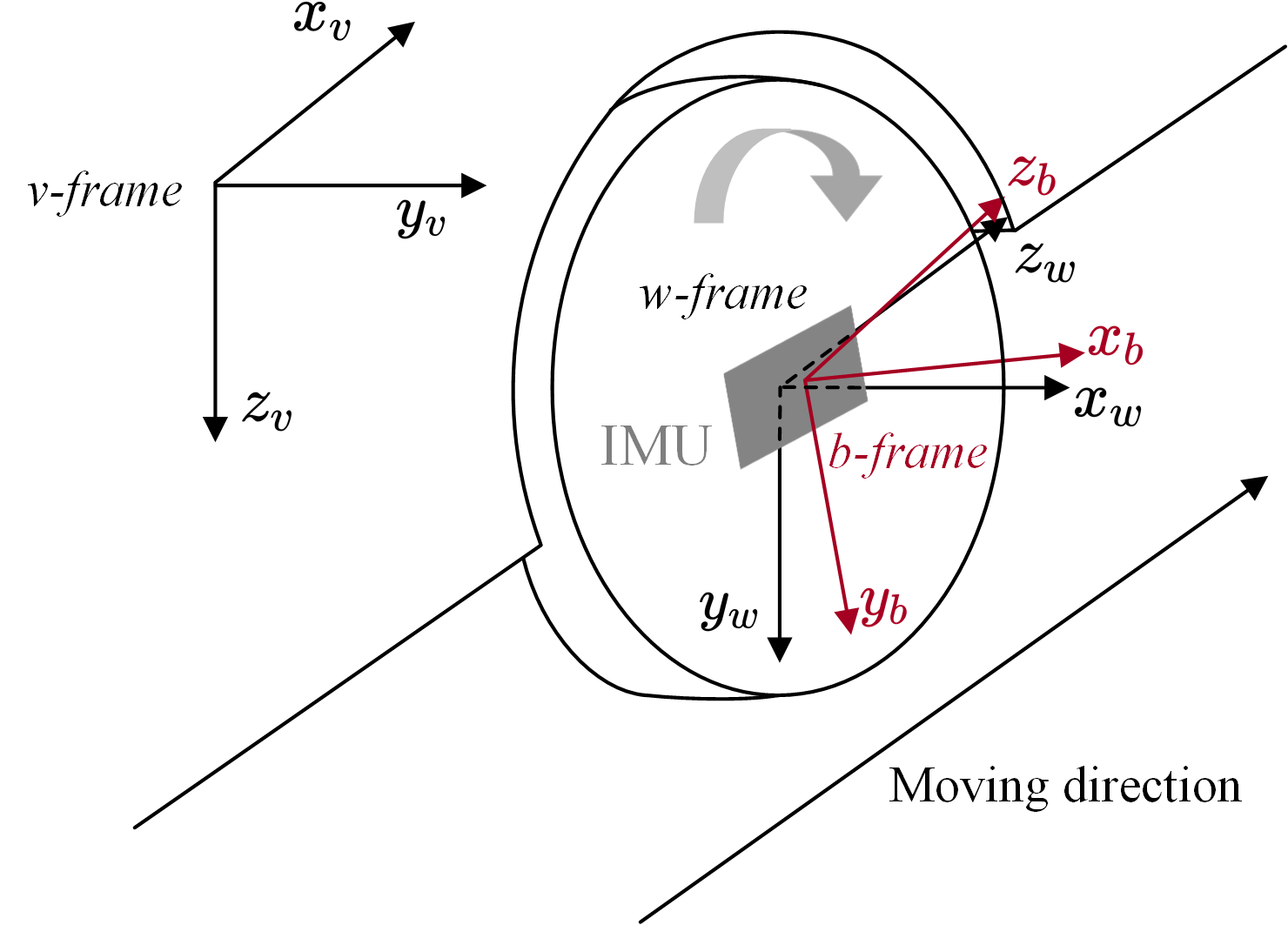}
	\caption{Installation scheme of Wheel-IMU and the definitions of the vehicle frame (\textit{v}-frame), wheel frame (\textit{w}-frame), and IMU body frame (\textit{b}-frame). The position and attitude misalignment errors between the \textit{b}-frame and the \textit{w}-frame are also depicted.}
	\label{fig2}
\end{figure}

Fig. \ref{fig2} illustrates the installation of Wheel-IMU and the definition of the related coordinate systems. To make the DR system indicate the vehicle state intuitively without being affected by vehicle maneuvers, the IMU has to be placed on a non-steering wheel of the vehicle. The \textit{v}-frame denotes the vehicle coordinate system, with the \textit{x}-axis pointing to the advancement direction of the host vehicle, \textit{z}-axis pointing down, and \textit{y}-axis directing right to complete a right-handed orthogonal frame, i.e., the forward-right-down system. The origin of the \textit{v}-frame is usually set at the vehicle mass center. The \textit{w}-frame denotes the wheel coordinate system. Its origin is at the rotation center of the wheel. Its \textit{x}-axis points to the right of the vehicle, and its \textit{y}- and \textit{z}-axes are parallel to the wheel surface to complete a right-handed orthogonal frame. The \textit{b}-frame denotes the IMU coordinate system, in which the accelerations and angular rates generated by the strapdown accelerometers and gyroscopes are resolved \cite{scherzinger1994}. The \textit{b}-frame axes are the same as the IMU's body axes. The \textit{x}-axis of the Wheel-IMU is aligned to the wheel rotation axis, pointing to the right of the vehicle, so as to avoid the singularity of the pitch angle ($\pm90^\circ$). Therefore, it can be considered that there only exists a periodic rolling angle between the \textit{w}-frame and the \textit{v}-frame given a stable vehicle structure. At the same time, the heading difference between the Wheel-IMU and the vehicle can be approximated as fixed (equal to 90$^\circ$), namely,
\begin{equation}
\psi_{b}^{n} = \psi_{v}^{n} + \pi/2
\label{headingdifference}
\end{equation}
where $\psi_{b}^{n}$ and $\psi_{v}^{n}$ denote the IMU heading and vehicle heading, respectively. \textit{n} indicates the \textit{n}-frame, which is a local-level frame with the origin coinciding with the \textit{b}-frame, \textit{x}-axis directing at the geodetic north, \textit{y}-axis east, and \textit{z}-axis downward vertically, i.e., the north-east-down system.

\subsection{Rotation of Wheel-IMU}
With the movement of the vehicle, Wheel-IMU continuously rotates around the rotation axis of the wheel, i.e., the $x_w$ axis in Fig. \ref{fig2}. Without loss of generality, we assume that the \textit{x}-axis of Wheel-IMU coincides with that of the \textit{w}-frame, pointing to the north and the gyroscope bias and the rotation angular rate of the wheel remain constant in one revolution. The constant measurement errors of the gyroscope of Wheel-IMU projected onto the \textit{n}-frame can be written as

\begin{equation}
\begin{aligned}
{\textbf{C}}_{b}^{n}\delta{{\bm{\varepsilon}}}^{b} &= \!\begin{bmatrix}
1 &\! \!0 &\!0\!\\ 
0 &\! \!\mathrm{cos}{\omega{t}} &\!-\mathrm{sin}{\omega{t}}\!\\ 
0 &\! \!\mathrm{sin}{\omega{t}} &\!\mathrm{cos}{\omega{t}}\!
\end{bmatrix} \begin{bmatrix}
\varepsilon_x\!\\ 
\varepsilon_y\!\\ 
\varepsilon_z\!
\end{bmatrix} \\
&= \begin{bmatrix}
\varepsilon_x\!\\ 
\varepsilon_{y}\mathrm{cos}{\omega{t}} - \varepsilon_{z}\mathrm{sin}{\omega{t}}\!\\ 
\varepsilon_{y}\mathrm{sin}{\omega{t}} + \varepsilon_{z}\mathrm{cos}{\omega{t}}\!
\end{bmatrix}
\end{aligned}
\label{rotationmodulation}
\end{equation}
where $\omega$ is the IMU rotation angular rate; ${\textbf{C}}_{b}^{n}$ is the rotation matrix from the \textit{b}-frame to the \textit{n}-frame; $\delta{{\bm{\varepsilon}}}^{b} = (\varepsilon_x, \varepsilon_y, \varepsilon_z )^\mathrm{T}$ are the constant gyroscope errors in three axes. The IMU attitude error caused by the gyroscope error can be obtained by integrating Eq. \ref{rotationmodulation}, i.e.,
\begin{equation}
{\int}_{\textit{Tc}}{\textbf{C}}_{b}^{n}\delta{{\bm{\varepsilon}}}^{b} = \!\begin{bmatrix}
{\int}_{\textit{Tc}} {\varepsilon_x}\!\\
{\int}_{\textit{Tc}} {\varepsilon_{y}\mathrm{cos}{\omega{t}} - \varepsilon_{z}\mathrm{sin}{\omega{t}}}\!\\
{\int}_{\textit{Tc}} {\varepsilon_{y}\mathrm{sin}{\omega{t}} + \varepsilon_{z}\mathrm{cos}{\omega{t}}}
\end{bmatrix}
\end{equation}
where $\textit{Tc}$ denotes the rotation period of the wheel. These equations show that because of the wheel rotation, the constant measurement error in the non-rotating axes of the gyroscope are modulated into sine wave in the \textit{n}-frame. After an integral period, the accumulated pitch and heading error are canceled. The IMU velocity errors caused by the accelerometer errors can be analyzed in a similar way. More detailed explanation of the rotation modulation can be found in \cite{du2015, Rahim2012}. 

In conclusion, by rotating the IMU around a fixed axis, the IMU measurement errors in the other two axes of the Wheel-IMU can be modulated into sinusoid signals. Thus, the velocity errors in the east and down directions, as well as the pitch and heading error caused by the IMU measurement errors can be eliminated. However, if the Wheel-IMU \textit{x}-axis is not parallel to the rotation axis, the rotation modulation would be dysfunctional.

\subsection{Misalignment Errors}

In Wheel-INS, the observations are from the \textit{v}-frame, including the vehicle speed and the NHC. Fusing the measurements in the vehicle frame with INS requires the knowledge of the installation relationship between the vehicle and Wheel-IMU. However, the misalignment problem is inevitable in any real system combining information from more than one sensor, which, if not handled properly, could cripple the system performance \cite{wu2009}. 

As illustrated in Fig. \ref{fig2}, the misalignment errors include the position misalignment (called ``lever arm"), pointing from the IMU center to the wheel rotation center expressed in the \textit{b}-frame, and the attitude misalignment (called ``mounting angles"), from the \textit{b}-frame to the \textit{w}-frame. The lever arm not only introduces potential errors to the projection of IMU velocity from the \textit{b}-frame to the \textit{v}-frame (cf. Eq. \ref{insvspeed}), but also brings centripetal acceleration to the accelerometer measurements, which would cause significant error in the DR system, especially when the vehicle moves at high speed. In practice, the lever arm can be determined by averaging multiple manual measurements. If the accuracy of the averaged results cannot meet the requirements, it can also be augmented into the state vector to be estimated online. 

The attitude misalignment can be described as a set of Euler angles, $\delta\phi$, $\delta\theta$, and $\delta\psi$. It can not only weaken the effect of the rotation modulation, but also cause incomplete projection of the real rotation angular rate on the \textit{x}-axis of the gyroscope. Furthermore, Eq. \ref{headingdifference} holds only when the mounting angles are effectively compensated. In addition, the attitude misalignment would introduce a periodic difference between the calculated vehicle heading and the real value. Because the roll mounting angle has no impact in Wheel-INS, we only take into consideration the pitch and heading misalignment angles. Note that the attitude misalignment problem in Wheel-INS is similar to that of IMU-based pipeline inspection gauges (PIGs) \cite{chen2019, chen2020}. The authors in \cite{chen2020} proposed a complete method to calibrate the mounting angles between the PIG frame and the IMU body frame, in which details of the calibration procedure and error analysis can be found. In our experiments (cf. Section V), we used this approach to calibrate and compensate the mounting angles before data processing.

\section{Methodology}
Although the constant bias of Wheel-IMU in the axes perpendicular to the rotation axis can be canceled by rotation, errors in the rotation axis and other types of random error still lead to rapid position drift in INS. Therefore, other means are required to limit the error accumulation.

Vehicle motion information (including the forward velocity and NHC) is a classical measurement to fuse with INS for ego-motion estimation of wheeled robots. In this study, the gyroscope measurements of Wheel-IMU and the wheel radius are used to calculate the wheel velocity without the requirement to install an external odometer or access to the onboard encoder of the vehicle. Additionally, the error-state EKF is employed to implement the information fusion. It is a widely used approach which recasts the problem of state estimation from the state domain to error-state domain, generating a linear state model \cite{Roumeliotis1999, Shin2005, Madyastha2011}. In this section, the state model and the observation model of Wheel-INS are described.
\subsection{Error State Model}
In Wheel-INS, conventional strapdown inertial navigation system is utilized to predict the vehicle state. The kinematic equations are described at length in the literature \cite{Shin2005, Groves2013, britting2010}; thus, we do not go into details here. For simplicity and efficiency, we use the error-state Kalman filter to mitigate the nonlinearity problem.

Because of the rotation of the wheel, the \textit{y}-axis and \textit{z}-axis of the inertial sensor change their directions around the \textit{x}-axis periodically; thus, it is difficult for the filter to distinguish and estimate the sensor errors in these two axes effectively. The scale factor error of the gyroscope in the \textit{x}-axis can converge soon in the system, as it can cause obvious error with the rotation of the wheel, whereas the gyroscope scale factor errors in the \textit{y}-axis and \textit{z}-axis are unobservable. Furthermore, the scale factor error of acceleration in the \textit{y}-axis and \textit{z}-axis can also be estimated jointly because they sense the gravity alternatively, and the horizontal motion assumption of the vehicle can make the errors evident. 

Although only estimating the observable states may reduce some computation cost, the cross-coupling errors of the inertial sensor triads couple the errors in the three axis together, especially for MEMS sensors; thus, the system performance degrades if we simply remove the unobservable sensor errors from the state vector, e.g., only estimating the scale factor error of the gyroscope in the \textit{x}-axis, not that in the \textit{y}- and \textit{z}- axes. A performance comparison of Wheel-INS with different dimensions of state vector is presented in Section \uppercase\expandafter{\romannumeral4}-B. The results indicate that the 21 dimensional state vector has better positioning performance. Therefore, the 21-state is employed in Wheel-INS, including three dimensional position errors, three dimensional velocity errors, attitude errors, residual bias errors and scale factor errors of the gyroscope and accelerometer. The state vector can be written as

\begin{equation}
\bm{x}(t)=\left[\left(\delta \bm{r}^{n}\right)^{\mathrm{T}} \quad\left(\delta \bm{v}^{n}\right)^{\mathrm{T}} \quad \bm{\phi}^{\mathrm{T}} \quad \bm{b}_{g}^{\mathrm{T}} \quad \bm{b}_{a}^{\mathrm{T}} \quad \bm{s}_{g}^{\mathrm{T}} \quad \bm{s}_{a}^{\mathrm{T}} \right]^{\mathrm{T}}
\label{statevector}
\end{equation}
where $\delta$ indicates the uncertainty of the variables; $\delta \bm{r}^{n}$, $\delta \bm{v}^{n}$ and $\bm{\phi}$ indicate the position, velocity and attitude errors of INS, respectively; $\bm{b}_{g}$ and $\bm{b}_{a}$ denote the residual bias errors of the gyroscope and the accelerometer, respectively; $\bm{s}_{g}$ and $\bm{s}_{a}$ are the residual scale factor errors of the gyroscope and accelerometer, respectively. In this paper, we only focus on the dead reckoning performance of Wheel-INS, and the initial heading of Wheel-INS is set by the reference system in our experiments. Therefore, the heading error in the state vector can be regarded as accumulated heading error rather than the absolute error. 

Several models have been developed to describe the time-dependent behavior of these errors \cite{Shin2005}. We adopt the Phi-angle model here. As Wheel-INS is a local DR system based on low-cost MEMS sensor, we do not take the earth rotation and the variation of the \textit{n}-frame into consideration. Thus, a simplified error state model can be expressed as
\begin{equation}
\left\{
\begin{aligned}
\delta \dot{\bm{r}}^{n}&=\delta \bm{v}^{n} \\
\delta \dot{\bm{v}}^{n}&=-\bm{\phi} \times \bm{f}^{n} + \textbf{C}_{b}^{n}\delta\bm{f}_{b} \\
\dot{\bm{\phi}}&=-\textbf{C}_{b}^{n}\delta\bm{\omega}_{ib}^{b}
\end{aligned}
\right.
\label{systemnodel}
\end{equation}
where $\delta\bm{\omega}_{ib}^{b}$ and $\delta\bm{f}^{b}$ are the errors of the gyroscope and accelerometer measurements, respectively, which can be expressed as $\delta\bm{\omega}_{ib}^{b} = \bm{b}_{g} + diag(\bm{\omega}_{ib}^{b})\bm{s}_{g}$ and $\delta\bm{f}^{b} = \bm{b}_{a} + diag(\bm{f}^{b})\bm{s}_{a}$; $diag(\cdot)$ is the diagonal matrix form of a vector. We choose the first-order Gauss-Markov process \cite{maybeck1982, Brown1992} to model the sensor errors. The continuous-time model and discrete-time model are written as
\begin{equation}
\begin{aligned}
\dot{x} &= - \dfrac{1}{\mathrm{T}}x+w\\
x_{k+1} &= e^{-\Delta{t}_{k+1}/\mathrm{T}}x_k+w_k
\end{aligned}
\label{sensorerrmodel}
\end{equation}
where $x$ is the random variable; $\mathrm{T}$ is the correlation time of the process, and $w$ is the driving white noise.

The continuous-time dynamic model for the error state can be written as

\begin{equation}
\dot{\bm{x}}(t) = \textbf{F}(t)\bm{x}(t) + \textbf{G}(t)\bm{w}(t)
\label{dynamicmodel}
\end{equation}
where $\bm{x}(t)$ is shown as Eq. \ref{statevector}; $\textbf{F}(t)$ is the system transition matrix; $\textbf{G}(t)$ is the system noise distribution matrix; $\bm{w}(t) = \left[(\bm{w}_g)^\mathrm{T}  (\bm{w}_a)^\mathrm{T}  (\bm{w}_{bg})^\mathrm{T}  (\bm{w}_{ba})^\mathrm{T}  (\bm{w}_{sg})^\mathrm{T}  (\bm{w}_{sa})^\mathrm{T}\right]^\mathrm{T}$ is the system noise, including the noises of the gyroscope, accelerometer, gyroscope bias error, accelerometer bias error, gyroscope scale factor error and  accelerometer scale factor error.  Details about matrices $\textbf{F}(t)$ and $\textbf{G}(t)$ can be found in Appendix. It can be observed that the system matrices are sparse which render the computational efficiency of Wheel-INS extremely high. 

It is worth mentioning that the scale factor error of the wheel may degrade the performance of Wheel-INS in some cases, and it cannot be substituted or assimilated by the errors of Wheel-IMU because they impact the navigation results in different ways. The scale factor error of the wheel should be estimated online if other absolute positioning information is available, such as GNSS.

\subsection{Observation Model}
The forward wheel velocity calculated by the gyroscope readings of Wheel-IMU and wheel radius can be written as
\begin{equation}
\begin{aligned}
\widetilde{v}^{v}_{wheel} &=\widetilde{\omega}_{x}r-e_v = ({\omega}_{x}+\delta{\omega}_{x})r-e_v\\ &={v}^{v}_{wheel}+r\delta{\omega}_x - e_v
\end{aligned}
\end{equation}
where $\widetilde{v}^{v}_{wheel}$ and ${v}^{v}_{wheel}$ are the observed and true vehicle forward velocity, respectively; $\widetilde{\omega}_{x}$ is the gyroscope output in the \textit{x}-axis; ${\omega}_{x}$ is the true value of the angular rate in the \textit{x}-axis of Wheel-IMU; $\delta{\omega}_{x}$ is the gyroscope measurement error; $r$ is the wheel radius, and $e_v$ is the observation noise, modeled as Gaussian white noise. 

The motion of the wheeled robots is generally governed by two non-holonomic constraints \cite{wu2009, Siegwart2011}, which refers to the fact that the velocity of the wheeled vehicle in the plane perpendicular to the forward direction is almost zero when the vehicle does not slide on the ground or jump off the ground \cite{Shin2005, sukkarieh2000}. By integrating with NHC, the 3D velocity observation can be expressed as
\begin{equation}
\widetilde{\bm{v}}^{v}_{wheel} =\begin{bmatrix}
\widetilde{v}^{v}_{wheel} &\! \!0 &\!0
\end{bmatrix}^\mathrm{T}-\bm{e}_v
\end{equation}
Because Wheel-IMU rotates with the wheel, the roll angle of the Wheel-IMU with respect to the \textit{v}-frame changes periodically. In consequent, the pitch angle of the vehicle cannot be obtained with the Wheel-IMU alone. In other words, it cannot be determined whether the vehicle is moving uphill or downhill by Wheel-INS. Therefore, we must assume that the vehicle is moving on a horizontal surface. Actually, this assumption does not cause significant error to the DR system, as illustrated by the car experiments in Section IV. According to Eq. \ref{headingdifference}, the Euler angles of the vehicle can be represented as 
\begin{equation}
\bm{\varphi}_{v}^{n} = \begin{bmatrix}
\phi_v^n\\
\theta_v^n\\
\psi_v^n
\end{bmatrix} = \begin{bmatrix}
0\\
0\\
\psi_b^n - \pi/2
\end{bmatrix}
\label{vehicleeulerangle}
\end{equation}
where $\bm{\varphi}_{v}^{n}$ is the attitude of the vehicle with respect to the \textit{n}-frame; $\phi$, $\theta$, and $\psi$ are the roll, pitch, and heading angle of the vehicle, respectively. The corresponding rotation matrix can be calculated by
\begin{equation}
\textbf{C}_{n}^{v} = \begin{bmatrix}
\mathrm{cos}\psi_v^n &\! -\mathrm{sin}\psi_v^n &\! 0\\
\mathrm{sin}\psi_v^n &\! \mathrm{cos}\psi_v^n &\! 0\\
0 &\! 0 &\! 1
\end{bmatrix}
\label{calCnv} 
\end{equation}

By performing the perturbation analysis, the INS-indicated velocity in the \textit{v}-frame can be written as
\begin{equation}
\begin{aligned}
\hat{\bm{v}}^v_{wheel} &= \hat{\textbf{C}}^v_n \hat{\bm{v}}^n_{IMU}\!+\!\hat{\textbf{C}}^v_n \hat{\textbf{C}}^n_{b} \left( \hat{\bm{\omega}}^{b}_{nb} \times \right) \bm{l}^{b}_{wheel}\\
&\approx {\textbf{C}}^v_n (\textbf{I} \!+\! \delta{\psi} \times)({\bm{v}}^n_{IMU} \!+\! \delta{\bm{v}}^n_{IMU})\\
&\quad +\! {\textbf{C}}^v_n(\textbf{I} \!+\! \delta\bm{\psi} \times)(\textbf{I} - \delta\bm{\phi} \times) \textbf{C}^n_b ({\bm{\omega}}^{b}_{nb}\!\times \!+\! \delta{\bm{\omega}}^{b}_{nb} \times) \bm{l}^{b}_{wheel}\\
&\approx {\bm{v}}^v_{wheel} \!+\! {\textbf{C}}^v_n \delta{\bm{v}}^n_{IMU} \!+\! {\textbf{C}}^v_n \left[ ({\textbf{C}}^n_b ({\bm{\omega}}^{b}_{nb} \times) \bm{l}^{b}_{wheel}) \times \right] \bm{\phi}\\
&\quad -\! {\textbf{C}}^v_n \left[ ({\bm{v}}^n_{IMU} \times) \!+\! ({\textbf{C}}^n_b ({\bm{\omega}}^{b}_{nb} \times) \bm{l}^{b}_{wheel}) \times \right] \delta\bm{\psi}\\
&\quad -\! {\textbf{C}}^v_n {\textbf{C}}^n_b(\bm{l}^{b}_{wheel} \times)\delta{\bm{\omega}}^{b}_{ib}
\end{aligned}
\label{insvspeed}
\end{equation} 
where $\textbf{C}_{n}^{v}$ is the rotation matrix from the \textit{n}-frame to the \textit{v}-frame, which can be transformed from the Euler angles $\bm{\varphi}_{v}^{n}$ by Eq. \ref{calCnv}; $\bm{\omega}_{nb}^{b}$ is the angular rate vector of the \textit{b}-frame with respect to the \textit{n}-frame projected to the \textit{b}-frame; $\bm{l}_{wheel}^b$ indicates the lever arm vector from the Wheel-IMU to the wheel center projected in the \textit{b}-frame; $\delta\bm{\psi}$ is the attitude error of the vehicle, which is only related to the heading error in the state vector; thus, it can be written as $\delta\bm{\psi} = \begin{bmatrix}
0 &\! 0 &\! \delta{\psi}_b^n
\end{bmatrix}^\mathrm{T}$. Then, the velocity error observation equation in the \textit{v}-frame can be written as
\begin{equation}
\begin{aligned}
\delta\bm{z}_v &= \hat{\bm{v}}^v_{wheel} - \widetilde{\bm{v}}^v_{wheel} \\
&= {\textbf{C}}^v_n \delta\bm{v}^n_{IMU}\!+\!\textbf{C}^v_n \left[ ({\textbf{C}}^n_b ({\bm{\omega}}^{b}_{nb} \times) \bm{l}^{b}_{wheel}) \times \right] \bm{\phi}\\
&\quad -\! {\textbf{C}}^v_n \left[ ({\bm{v}}^n_{IMU} \times) \!+\! ({\textbf{C}}^n_b ({\bm{\omega}}^{b}_{nb} \times) \bm{l}^{b}_{wheel}) \times \right] \delta\bm{\psi}\\
&\quad -\! {\textbf{C}}^v_n {\textbf{C}}^n_b(\bm{l}^{b}_{wheel} \times)\delta{\bm{\omega}}^{b}_{ib}
\end{aligned}
\end{equation}

Although the heading error can be limited effectively by the continuous rotation with the wheel, it accumulates when the vehicle remains stationary for a long time. In this case, zero-integrated heading rate measurements (ZIHRs) \cite{Shin2005} can be performed to correct the heading. It is worth mentioning that the Wheel-IMU is more sensitive to vehicle motion; hence, it can perceive the vehicle motion more accurately than the IMU placed on the vehicle body.

\section{Experimental Results}
This section presents real-world experimental results to illustrate the positioning performance of Wheel-INS and analyze its characteristics. Firstly, the  experimental conditions are described. Secondly, we compare the DR performance between Wheel-INS and ODO/INS. We also conduct experiments to illustrate the insensitivity to constant gyroscope bias of Wheel-INS. Finally, we discuss the key characteristics of Wheel-INS by well-designed tests, including the influence of the mounting angles and the comparison of the positioning results with different dimensions of the state vector.
\subsection{Experimental Description}
Field tests were conducted in three different scenarios in Wuhan City, China, using two different wheeled vehicles. One was the Pioneer 3DX robot, a typical differential wheeled robot, and the other was a car. The Pioneer robot was used for two trajectories and the car for one trajectory. Fig. 3 shows the experimental platforms.
\begin{figure}[htbp]
	\centering
	\subfigure[Pioneer 3DX robot.]{
		\includegraphics[width=8.6cm]{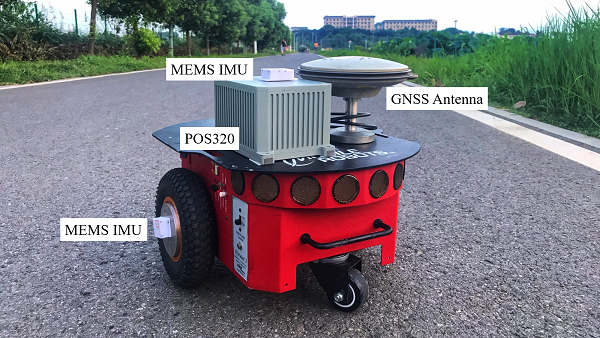}
	}
	\quad
	\subfigure[Pedestrain car.]{
		\includegraphics[width=8.6cm]{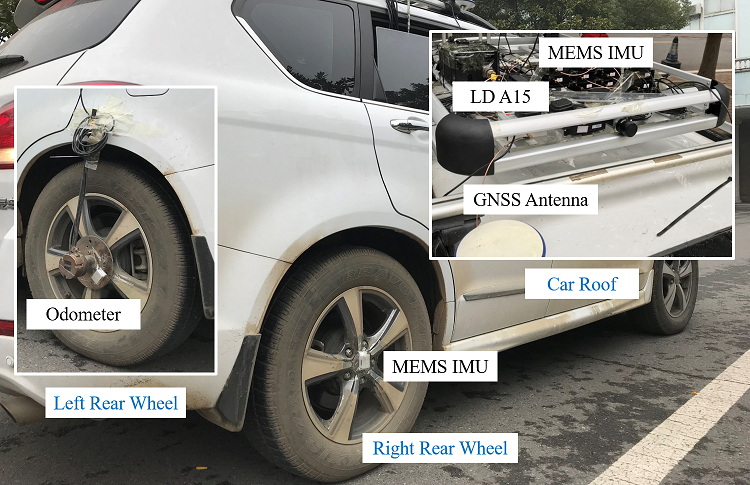}
	}
	\caption{Experimental platforms.}
	\label{fig3}
\end{figure}
The MEMS IMU used in the experiments was self-developed, including four ICM20602 inertial sensor chips, a chargeable battery module, a SD card for data collection, a micro processor, and a Bluetooth module for communication and data transmission. One can connect an android mobile phone to the MEMS IMU through Bluetooth communication to start and end the data collection. We collected the outputs of two chips in one trajectory as two sets of experimental data for post-processing. The MEMS IMUs were carefully placed on the wheel to make them as close as possible to the wheel center. To compare its performance with that of ODO/INS, a MEMS IMU of the same type was placed on the vehicle roof. The wheel odometer data was also recorded in the experiments. For the car, the wheel speed data came from an externally installed odometer. The root-mean-squared error (RMSE) of the wheel velocity measurements of the two vehicles were 0.03 m/s and 0.08 m/s, respectively. As shown in Fig. \ref{fig3}, the two vehicles were also equipped with high-accuracy IMUs to provide pose ground truth: a POS320 (MAP Space Time Navigation Technology Co., Ltd., China) for the robot experiments and a LD A15 (Leador Spatial Information Technology Co., Ltd., China) for the car experiments. Their main technique parameters are listed in Table \ref{Tabel1}. The reference data were processed through a smoothed post-processed kinematic (PPK)/INS integration method.

\begin{table}[h]
	\centering
	\caption{Technical Parameters of the IMUs Used in the Tests}
	\label{Tabel1}
	\begin{threeparttable}
		\begin{tabular}{m{1.2cm}<{\centering}m{1.3cm}<{\centering}m{1.3cm}<{\centering}m{1.3cm}<{\centering}m{1.3cm}<{\centering}}
			\toprule
			IMU & Gyro Bias\newline ($deg/h$) & {ARW\tnote{*} \newline ($deg/\sqrt{h}$)}  & {Acc.\tnote{*} Bias \newline ($m/s^2$)} & {VRW\tnote{*} \newline ($m/s/\sqrt{h}$)} \\
			\midrule
			\specialrule{0em}{3pt}{3pt}
			LD A15 & 0.02 & 0.003 & 0.00015 & 0.03 \\
			\specialrule{0em}{3pt}{3pt}
			POS320 & 0.5  & 0.05 & 0.00025 & 0.1\\
			\specialrule{0em}{3pt}{3pt}
			ICM20602 & 200 & 0.24 & 0.01 & 3\\
			\bottomrule
		\end{tabular}
		\begin{tablenotes}   
			\footnotesize            
			\item[*]ARW denotes the angle random walk; Acc. denotes the accelerometer; VRW denotes the velocity random walk.      
		\end{tablenotes}            
	\end{threeparttable}
\end{table}

\begin{figure}[htbp]
	\centering
	\subfigure[Track I: small-scale repeated trajectory in crossroads in the Information Department of Wuhan University.]{
		\includegraphics[width=8.6cm]{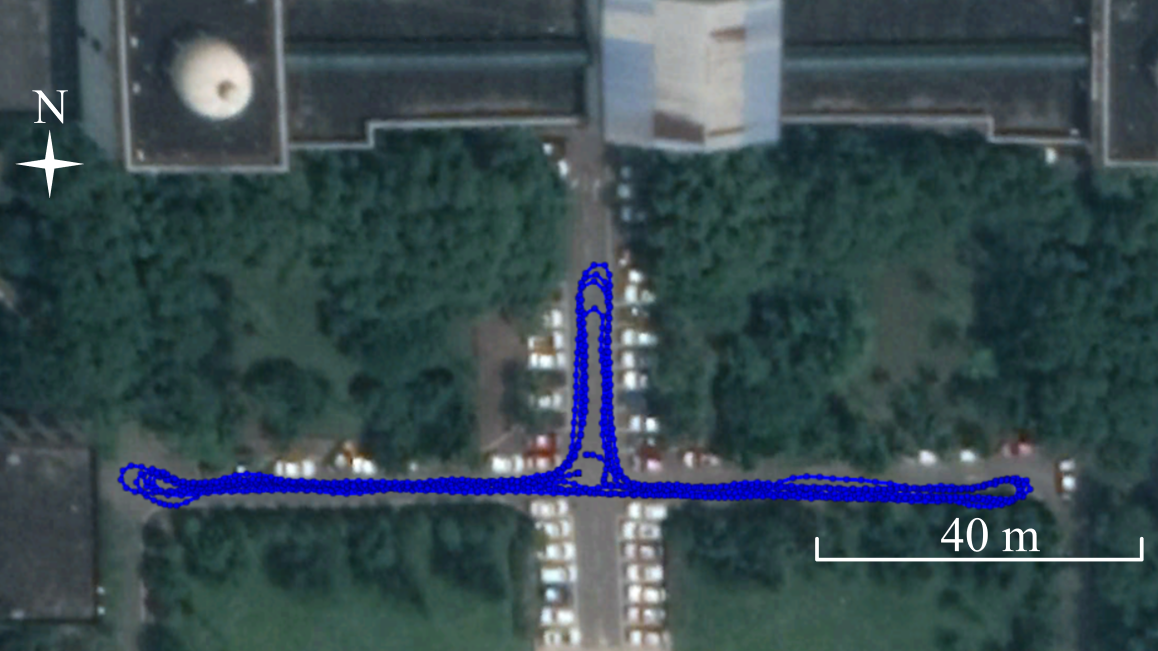}
	}
	\quad
	\subfigure[Track II: polyline trajectory with no return in experimental farms in Huazhong Agriculture University.]{
		\includegraphics[width=8.6cm]{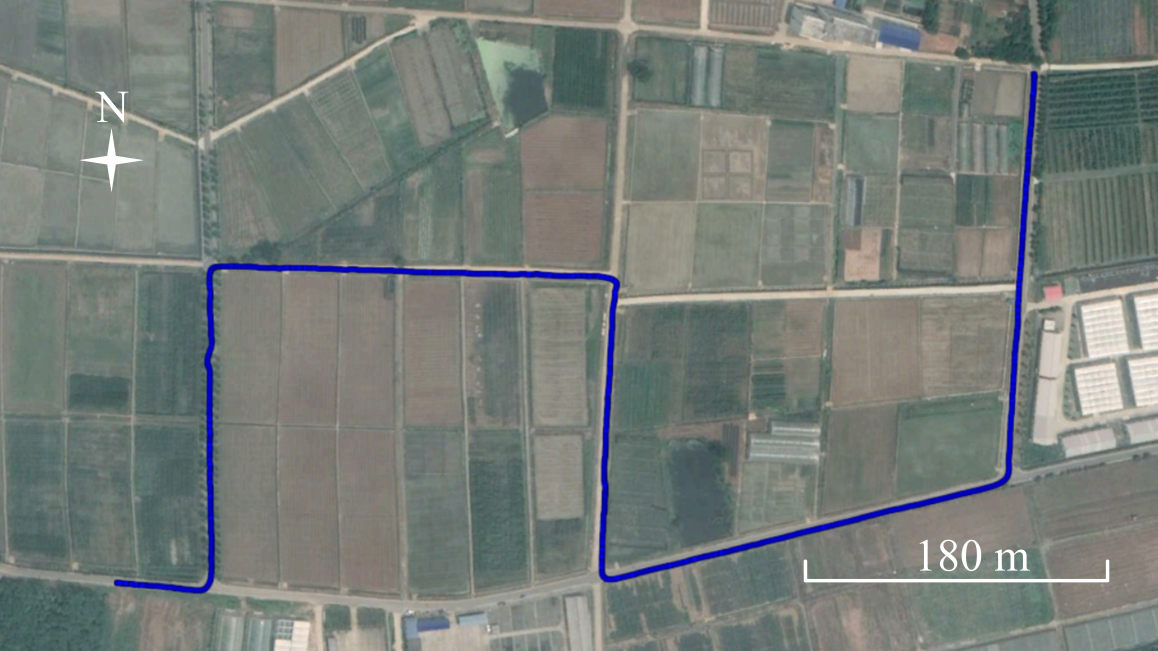}
	}
	\quad
	\subfigure[Track III: large-scale loop trajectory in Wuhan University campus]{
		\includegraphics[width=8.6cm]{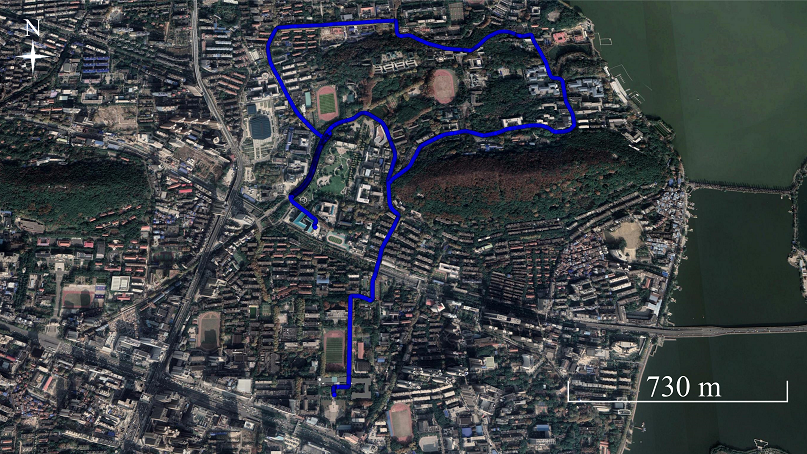}
	}
	\caption{Experimental trajectories.}
	\label{fig4}
\end{figure}

Fig. \ref{fig4} shows the three test trajectories. Track I is a loopback trajectory in a small-scale environment in the Information Department of Wuhan University, on which the robot moved for approximately five times. Track II is a polyline trajectory with no return in Huazhong Agriculture University. Track III is a large loop trajectory in the campus of Wuhan University, on which the car traveled for approximately two times. The vehicle motion information of all the six tests is presented in Table \ref{Tabel2}.

\begin{table}[h]
	\centering
	\caption{Vehicle Motion Information in the Experiments}
	\label{Tabel2}
	\begin{tabular}{ccccc}
		\toprule
 		{Test} & Track & Vehicle & \makecell{Average \\ speed ($m/s$)} & \makecell{Total \\ distance ($m$)} \\
		\midrule
		\specialrule{0em}{2pt}{2pt}
		\makecell{1 \\ 2 } & \makecell{I} & \multirow{3}*{Pioneer 3DX}  & \makecell{1.39 } & \makecell{$\approx$1227 }\\
		
		%\specialrule{0em}{2pt}{2pt}
		\makecell{3 \\ 4 } & \makecell{II} & & \makecell{1.25 } & \makecell{$\approx$1146 }\\
		%\cline{1-5}
		%\specialrule{0em}{2pt}{2pt}
		\makecell{5 \\ 6 } & \makecell{III} & \makecell{Car} & \makecell{4.70} & \makecell{$\approx$12199}\\
		\bottomrule
	\end{tabular}   
\end{table}

The initial heading and position of both Wheel-INS and ODO/INS were given by the reference system directly. On the one hand, the attitude misalignment between the reference IMU and the vehicle was calibrated and compensated in advance. On the other hand, we focused mainly on the DR performance of Wheel-INS. Therefore, the heading estimation error of the systems can be considered as the accumulated error without initial heading bias. However, other initial alignment methods should be investigated for practical applications because a reference system is not always available. Additionally, the static IMU data before the vehicle started moving were used to obtain the initial roll and pitch angle, as well as the initial value of the gyroscope bias. Other inertial sensor errors were set as zero. The initial states in ODO/INS were determined in the same way.

\begin{figure*}[htbp]
	\centering
	\subfigure[Estimated trajectories against ground truth in Test 1.]{
		\includegraphics[width=8.5cm]{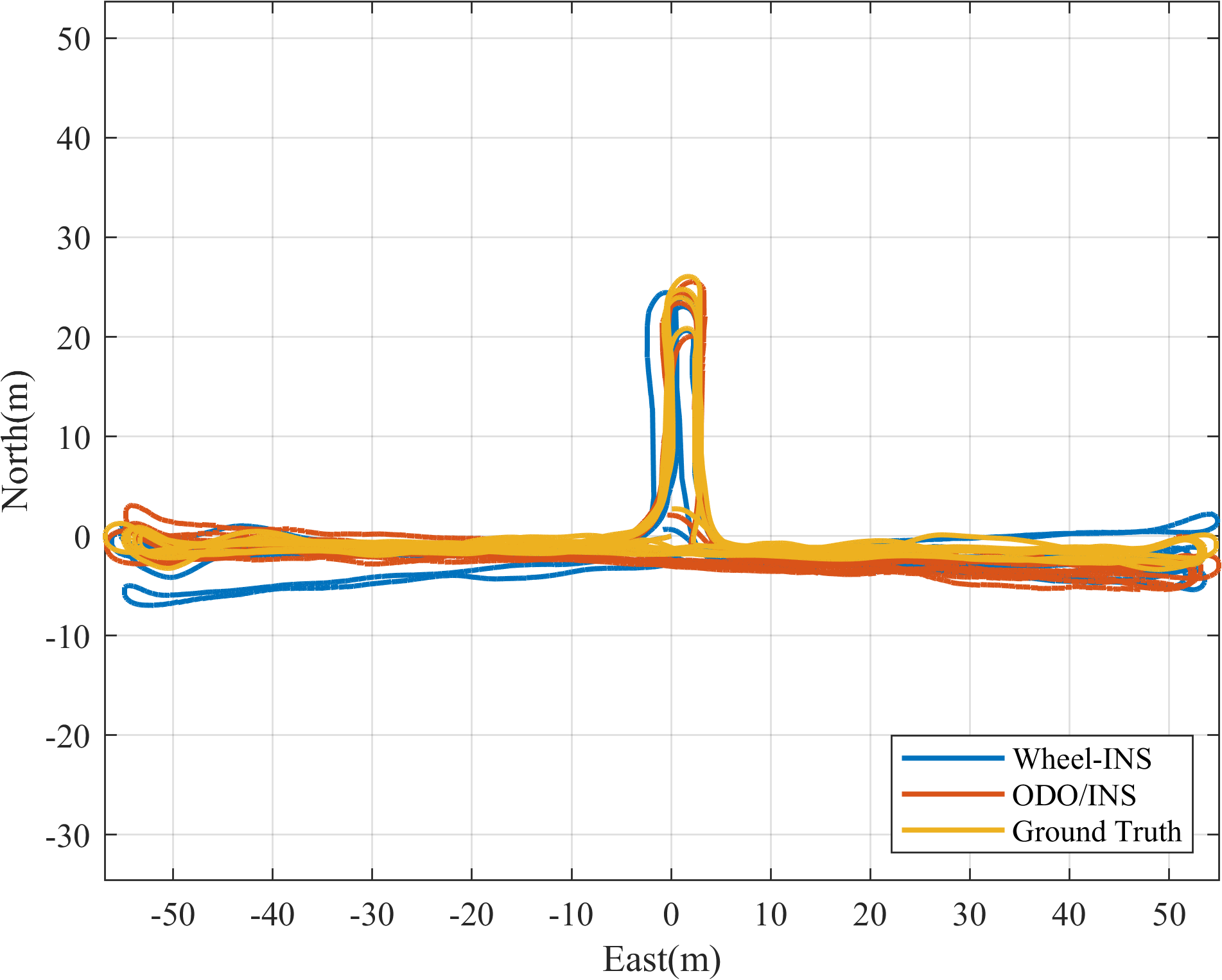}
	}
	\quad
	\subfigure[Positioning and heading errors in Test 1.]{
		\includegraphics[width=8.5cm]{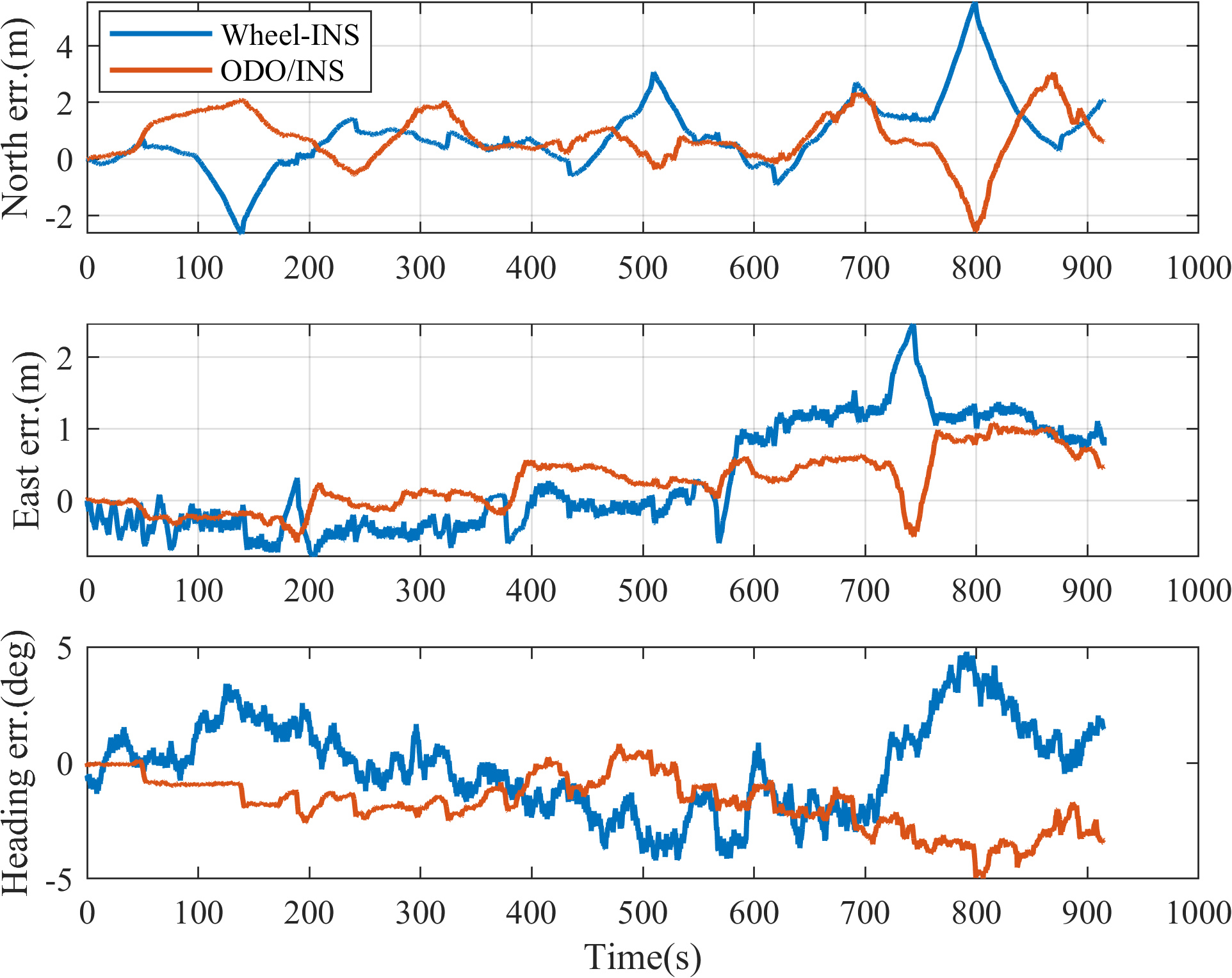}
	}
	\quad
	\subfigure[Estimated trajectories against ground truth in Test 3.]{
		\includegraphics[width=8.5cm]{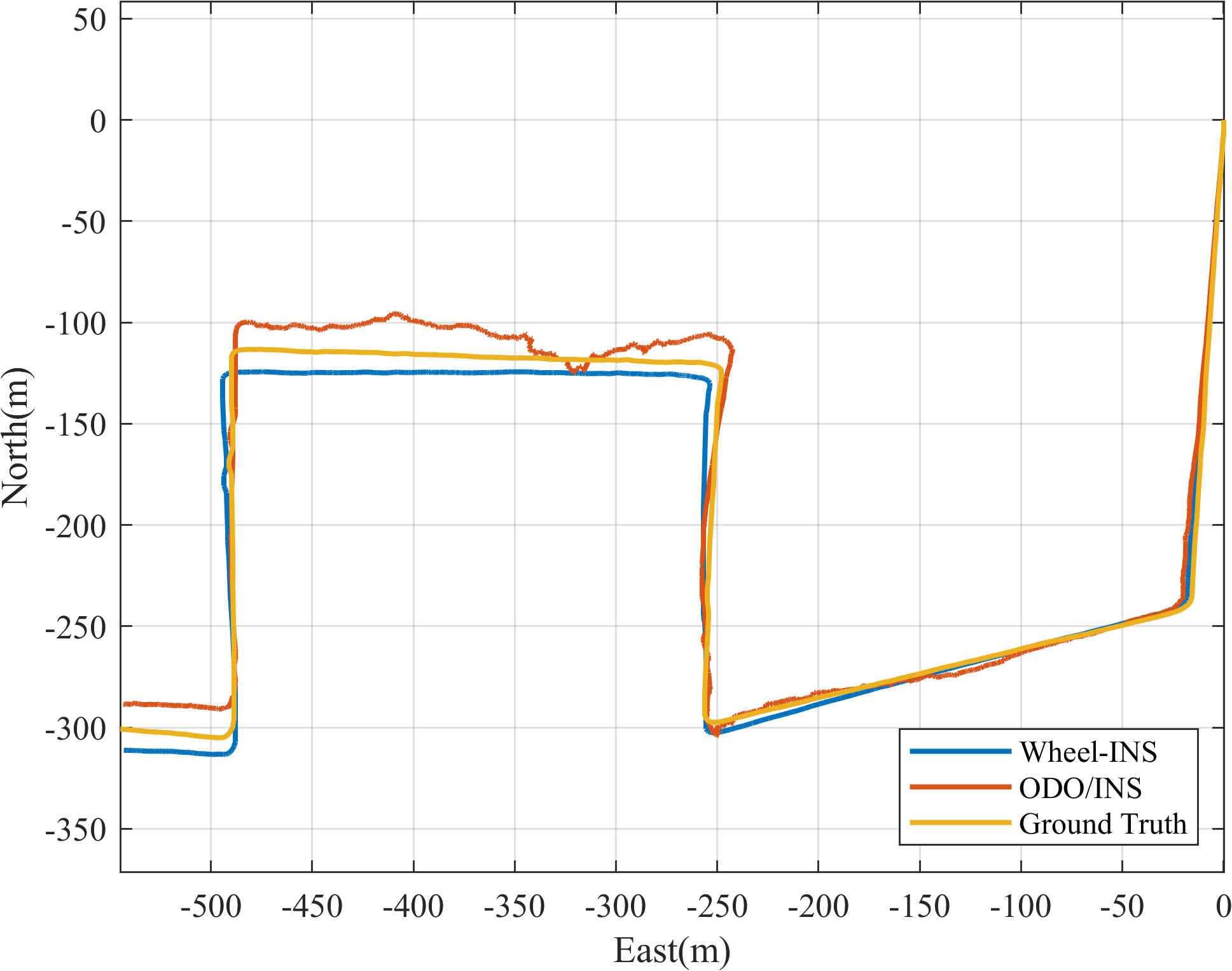}
	}
	\quad
	\subfigure[Positioning and heading errors in Test 3.]{
		\includegraphics[width=8.5cm]{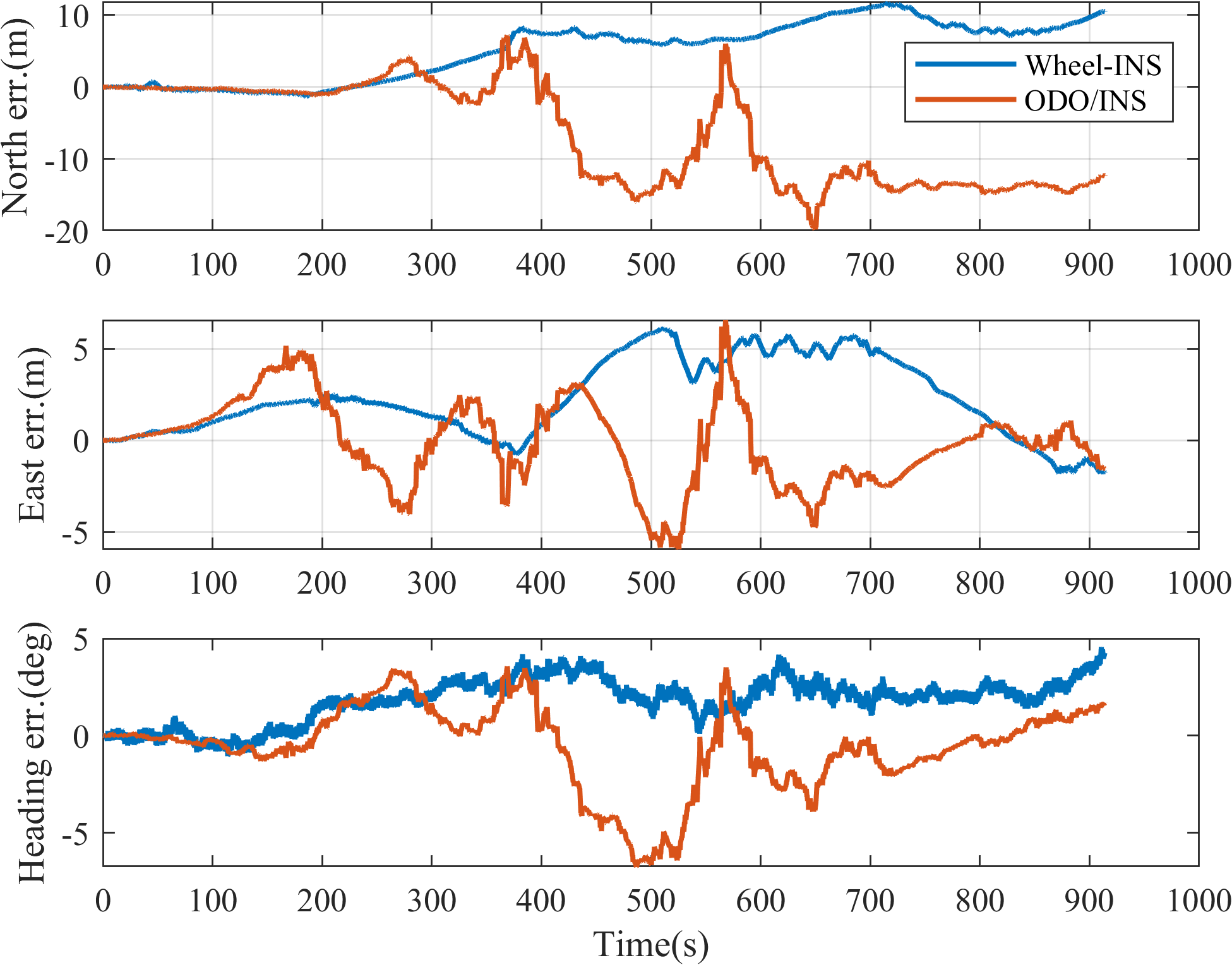}
	}
	\quad
	\subfigure[Estimated trajectories against ground truth in Test 5.]{
		\includegraphics[width=8.5cm]{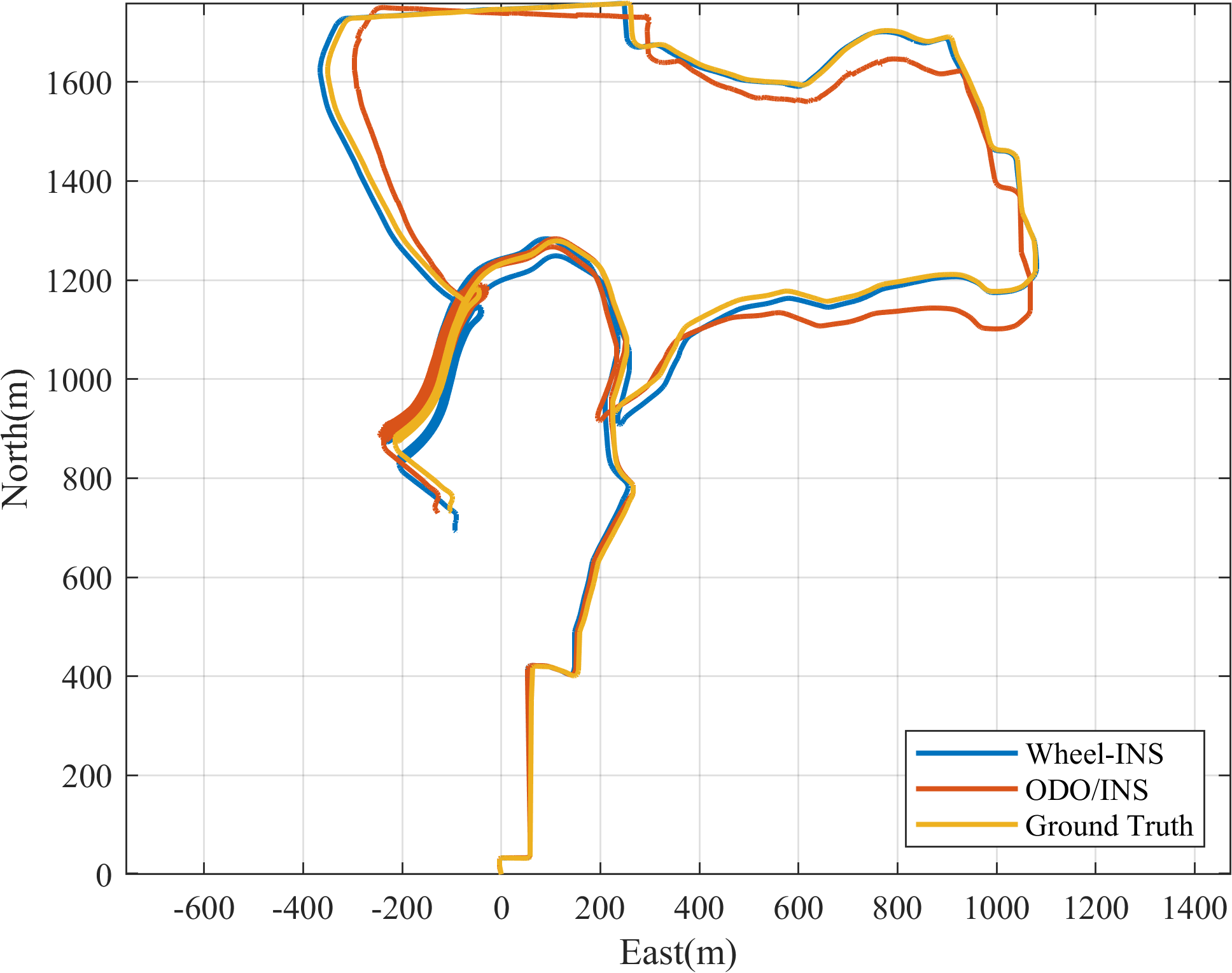}
	}
	\quad
	\subfigure[Positioning and heading errors in Test 5.]{
		\includegraphics[width=8.5cm]{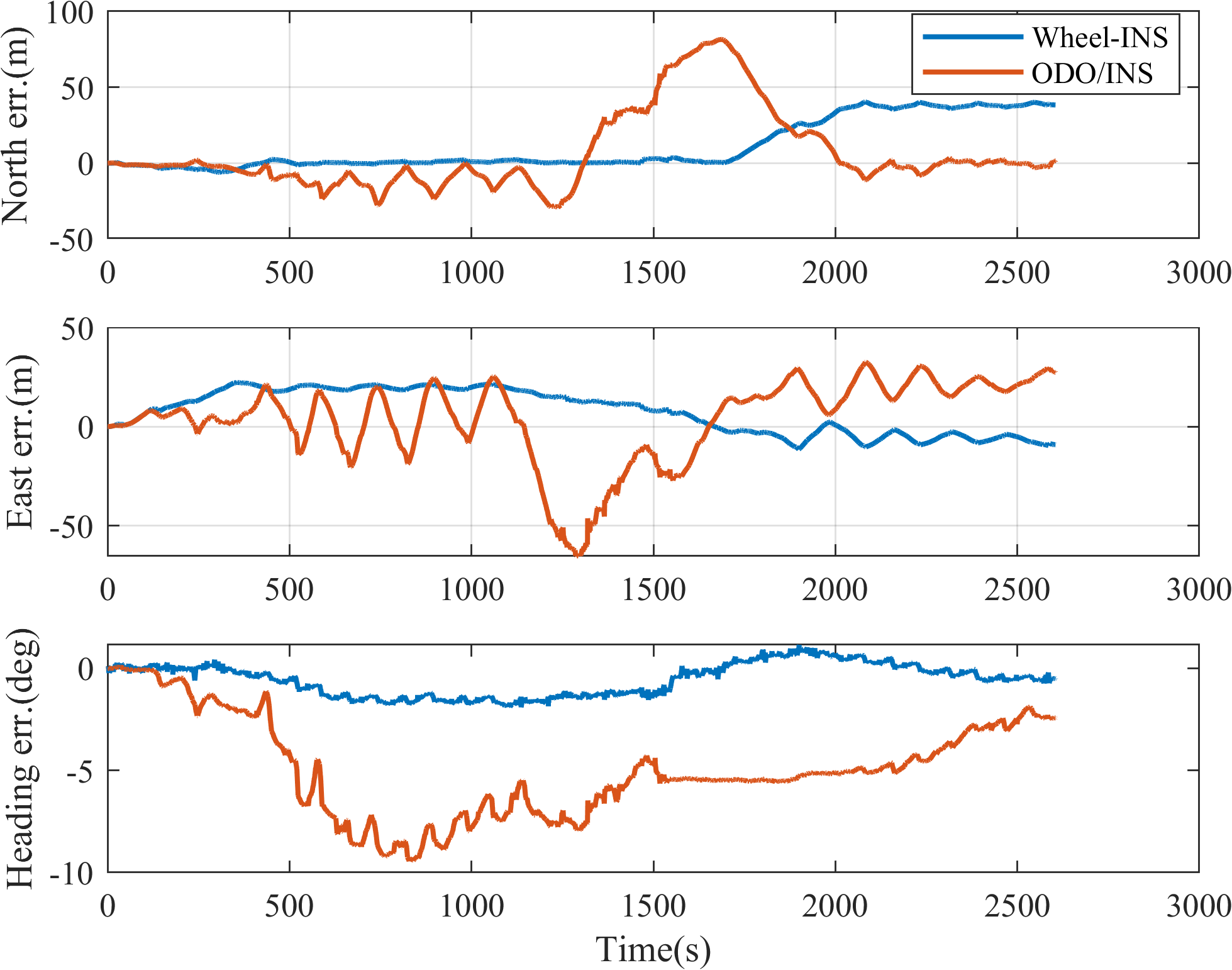}
	}
	\caption{Estimated trajectories and corresponding position and heading errors of Wheel-INS and ODO/INS in three experiments.}
	\label{fig5}
\end{figure*}

\subsection{Performance Comparison between Wheel-INS and ODO/INS}

\subsubsection{Comparison of Position and Heading Accuracy}
Fig. \ref{fig5} compares the positioning and heading errors of Wheel-INS and ODO/INS in Test 1, Test 3 and Test 5, respectively. We argue that calculating the misclosure error or the maximum position drift of the entire trajectory to demonstrate the positioning performance of a DR system is not optimal. The reason is that the loops in the trajectory can suppress the error accumulation to some extent thus make these methods not strict, especially for INS in which the position error usually drifts along one direction. It can be observed from Fig. \ref{fig5}(a) and (b) that when the robot turns around, the positioning error starts to drift along the opposite direction. Therefore, we adopted a different metric to evaluate the system performance. Firstly, we accumulated the moving distance of the robot by a certain increment ($l$). Then, we calculated the maximum horizontal position error drift rate within each distance ($l$, $2l$, $3l$, ...). Finally, the mean value (MEAN) and standard deviation (STD, $1\sigma$) of these segmented drift rates were computed as the indicator of positioning performance. This approach is similar to the odometry evaluation metric used in the KITTI dataset \cite{geiger2012}, but we segmented the trajectory only from the starting point. With regard to the heading error, the maximum (MAX) and RMSE were calculated as the evaluation criterion. In our experiments, we chose $l$ as 100 m. Table \ref{Table3} lists the error statistics of Wheel-INS and ODO/INS in all the six experiments.

\begin{table}[h]
	\centering
	\caption{Positioning and Heading Error Statistics of Wheel-INS and ODO/INS}
	\label{Table3}
	\begin{threeparttable}
		\begin{tabular}{p{0.7cm}<{\centering}p{1.4cm}<{\centering}p{1.1cm}<{\centering}p{1.1cm}<{\centering}p{0.8cm}<{\centering}p{0.8cm}<{\centering}}
			\toprule
			\multirow{2}*{\makecell{Test}} & \multirow{2}*{\makecell{System}} & \multicolumn{2}{c}{Position drift rate ($\%$)} & \multicolumn{2}{c}{Heading error $(^\circ)$} \\
			%\cline{3-6}
			\specialrule{0em}{1.5pt}{1.5pt}
			& & MEAN & STD & RMS & MAX \\
			\midrule
			\multirow{2}{*}{1}&{Wheel-INS}& 0.59& 0.30&\textbf{1.93}&4.79\\
			                  &{ODO/INS}  & \textbf{0.58}& 0.45&2.11&5.11\\
			\multirow{2}{*}{2}&{Wheel-INS}& 1.43& 0.54&3.88&7.93\\
			                  &{ODO/INS}  & \textbf{0.57}& 0.32&\textbf{2.82}&4.79\\
			\multirow{2}{*}{3}&{Wheel-INS}& \textbf{1.17}& 0.27&\textbf{2.16}&4.56\\
			                  &{ODO/INS}  & 2.25& 0.68&2.88&8.00\\
			\multirow{2}{*}{4}&{Wheel-INS}& \textbf{1.78}& 0.26&\textbf{4.44}&10.88\\
			                  &{ODO/INS}  & 2.20& 0.88&5.77&9.87\\
			\multirow{2}{*}{5}&{Wheel-INS}& \textbf{0.62}& 0.42&\textbf{0.96}&1.91\\
			                  &{ODO/INS}  & 1.14& 0.65&1.38&3.30\\
			\multirow{2}{*}{6}&{Wheel-INS}& \textbf{0.83}& 0.43&\textbf{1.60}&4.97\\
			                  &{ODO/INS}  & 1.62& 1.04&2.65&6.72\\
			\bottomrule
		\end{tabular}
		 
	\end{threeparttable}
\end{table}

From the comparison of positioning and heading performance between Wheel-INS and ODO/INS, the following information can be obtained.
\begin{itemize}

\item In all the six experiments, the horizontal position drift rates of Wheel-INS are between 0.50\% and 1.80\%, and the same figures of ODO/INS are between 0.50\% and 2.30\%. The average positioning and heading accuracy of Wheel-INS have been respectively improved by 23\% and 15\% comparing with ODO/INS.
\\
\item In most experiments, Wheel-INS outperforms ODO/INS in both terms of position and heading estimation accuracy. In Test 1, the positioning accuracy of the two systems are at a similar level while the heading estimation error of Wheel-INS is slightly smaller than that of ODO/INS. However, in Test 2, the mean position drift rate of ODO/INS is less than half of that of Wheel-INS. As we explained previously, we firstly calculated and compensated the gyroscope bias using the static IMU data before the vehicles started moving in our experiments. And during a short period of time (e.g., 15 min for Track I), the gyroscope bias could remain stable; thus it only caused small heading drift in ODO/INS. Therefore, the advantages of Wheel-INS comparing with ODO/INS are not obvious in Test 1 and Test 2.
%In addition, the U-turns in Track I also helped to limit the position error accumulation.
\\
\item In Track II (Test 3 and Test 4), there were many bumpy road sections. It can be observed from Fig. \ref{fig5}(c) and (d) that the DR errors of ODO/INS vibrate significantly while the errors of Wheel-INS are smoother. In this situation, the ground vehicles shook severely thereby broke the assumption of NHC. In addition, because the lever arm between the wheel center and the IMU attached on the vehicle body is larger than that in Wheel-INS, a more noticeable error would be generated in ODO/INS during the velocity projection when fusing the vehicle velocity with INS. As a result, the positioning error of ODO/INS would be larger and more unstable than that of Wheel-INS in bumpy roads. 
\\
\item In Track III (Test 5 and Test 6), the experiments were conducted using a car. It is obvious that the performance of Wheel-INS in these two tests are better than that in other experiments in which the wheeled robot was used as the test platform. The reason can be summarized as twofold. Firstly, the wheel and axle structure of a car is much stabler than that of a robot which provides a great condition for Wheel-INS. For example, the assumption of NHC is more convinced for cars. Secondly, the speed of the car was faster than the robot in our experiments (cf. Table \ref{Tabel2}). A relatively faster wheel speed would allow Wheel-INS gain more form the rotation modulation, resulting in higher accuracy in heading.
\\
\item Moreover, in Track III, the car moved along a large-scale trajectory with many continuous uphill and downhill roads (the maximum gradient was about 10$^\circ$), but the positioning accuracy of Wheel-INS is still considerable in Test 5 and Test 6. These results indicate that although the algorithm require the vehicle to move on a horizontal plane, it would not introduce significant error if there are some degrees of incline in the road. That is to say, Wheel-INS can be applied to most of the regular city roads.
\end{itemize}

\begin{figure*}[htbp]
	\centering
	\subfigure[Trajectories of Wheel-INS with different gyroscope bias in Test 1.]{
		\includegraphics[width=8.5cm]{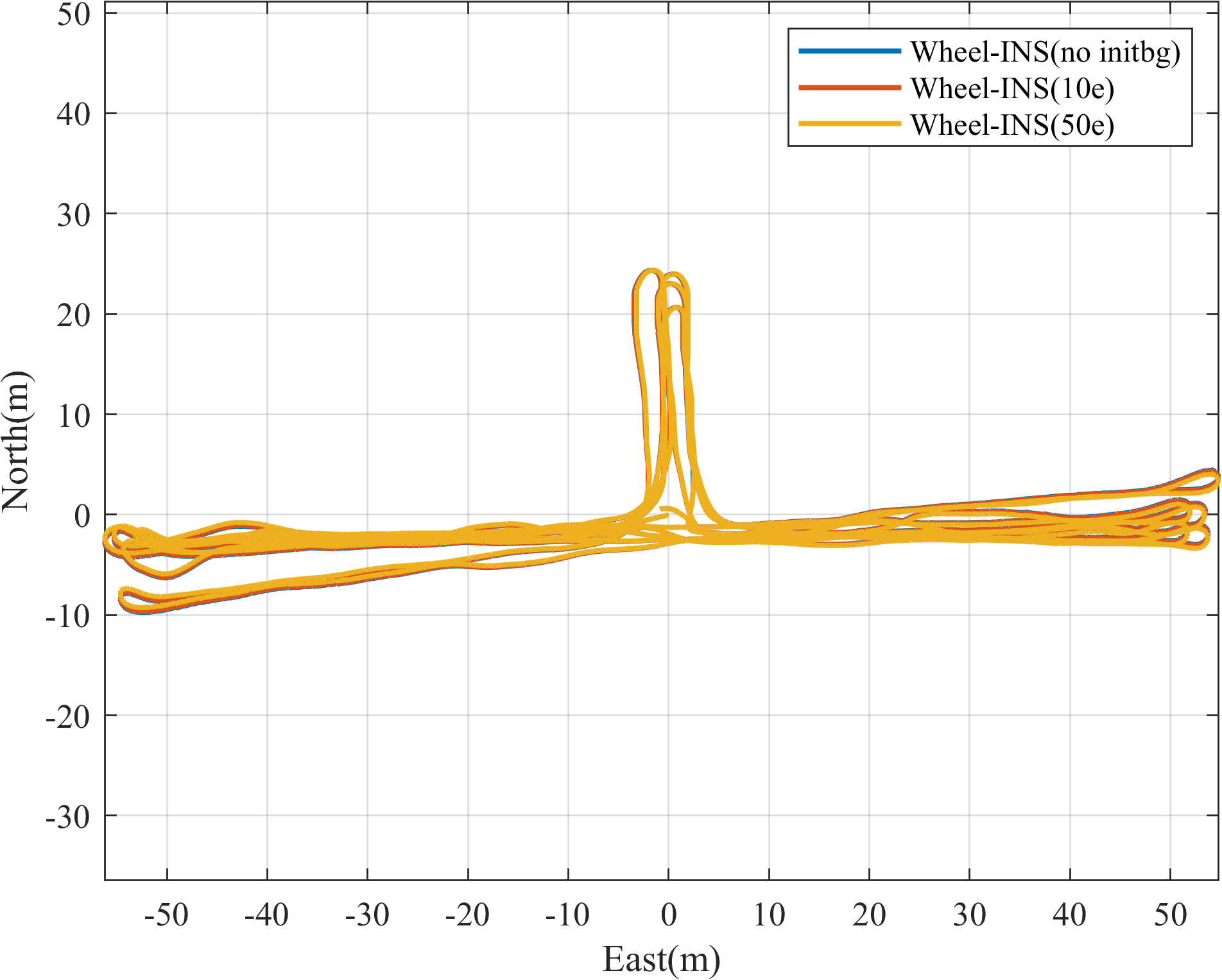}
	}
	\quad
	\subfigure[Trajectories of ODO/INS with different gyroscope bias in Test 1.]{
		\includegraphics[width=8.5cm]{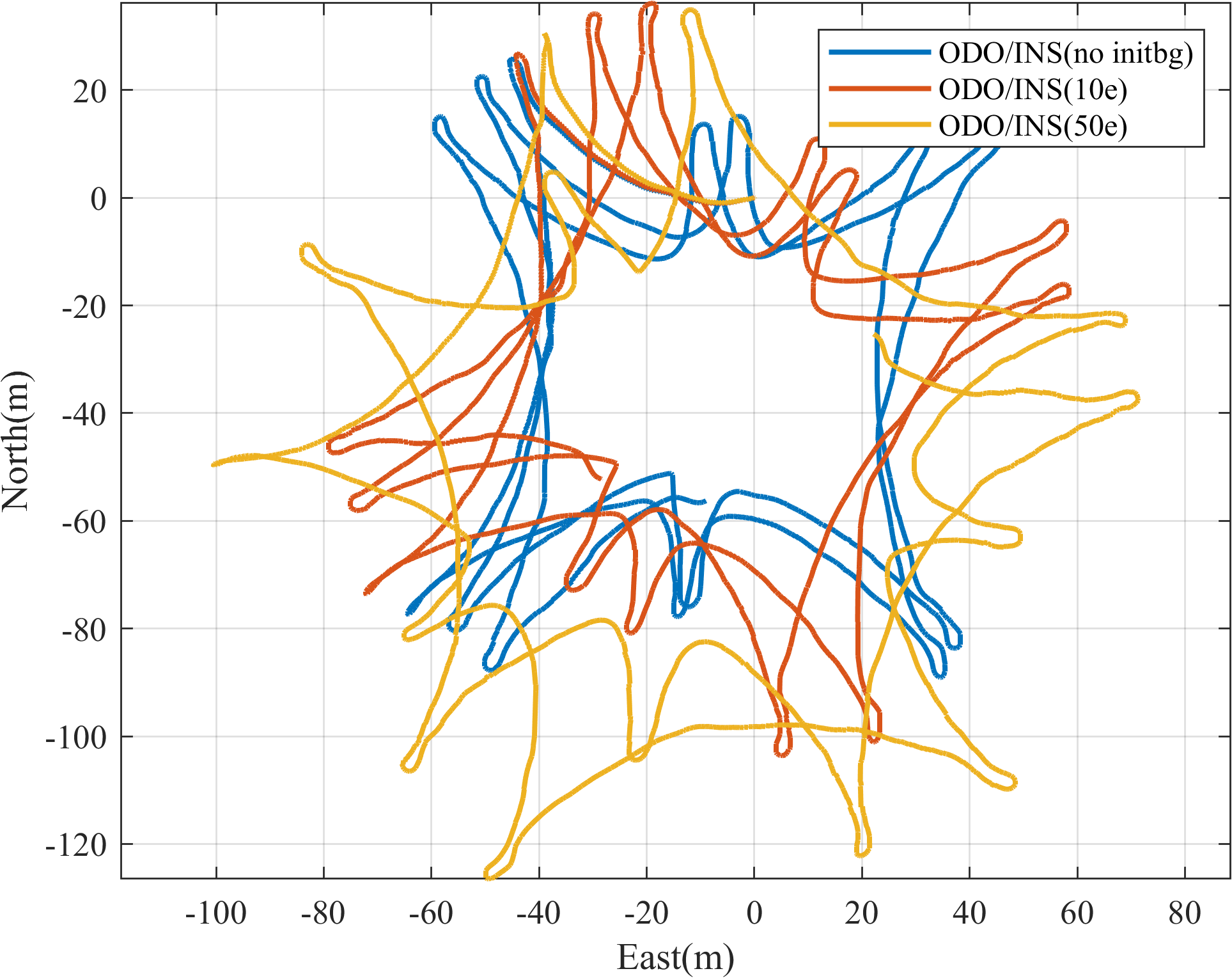}
	}
	\caption{Estimated trajectories of Wheel-INS and ODO/INS given different conditions on initial gyroscope bias.``no initbg'' indicates that the initial gyroscope bias was not compensated; ``10e'' indicates that ten times of earth rotation angular rate was manually added to the raw gyroscope data; ``50e'' indicates that fifty times of earth rotation angular velocity was added.}
	\label{fig6}
\end{figure*}

As for the velocity estimation, the mean RMS of the horizontal velocity error of Wheel-INS and ODO/INS in all the six experiments are 0.16 $m/s$ and 0.17 $m/s$, respectively. In both Wheel-INS and ODO/INS, the observations of the filter systems are the forward velocity of the vehicle and NHC. As there is no significant difference between the wheel speed provided by Wheel-IMU and the encoder, the velocity estimation accuracy of the two systems are at a similar level.

\subsubsection{Comparison of Sensitivity to Constant Gyroscope Bias}

To illustrate the rotation modulation effect in Wheel-INS, we conducted a set of comparative experiments of Test 1 given different level of gyroscope bias. Firstly, we did not estimated and compensated the initial gyroscope bias. Secondly, we intentionally add some constant errors onto the raw gyroscope data of Wheel-IMU and the IMU mounted on the robot body. Hereby ten times and fifty times of earth rotation angular rate ($\approx15^\circ/h$) were respectively added. Then we compare the positioning performance of Wheel-INS and ODO/INS under these conditions. As shown in Fig. \ref{fig6}, it is evident that the estimated trajectories of Wheel-INS change negligibly even ten and fifty times of earth rotation angular rate was added, while the trajectories of ODO/INS dramatically drift when the initial gyroscope bias error was not estimated in advance and it gets worse when more gyroscope bias was added. As the gyroscope bias cannot be effectively estimated by integrating vehicle motion measurements (forward velocity and NHC) in ODO/INS, it can induce large heading drift. However, the constant gyroscope error does not degrade the performance of Wheel-INS thanks to the rotation modulation. Therefore, Wheel-INS is significantly more insensitive to the constant gyroscope bias than ODO/INS.

Note that although zero-velocity updates (ZUPTs) \cite{dissanayake2001, Kilic2019} and ZIHRs also benefit to the estimation of gyroscope bias, they are merely random signals which cannot be relied on. In addition, mistakes in detecting the static time frame of the vehicle would instead undermine the stability of the system.

\subsection{Discussion on the Characteristics of Wheel-INS}
%In this section, we design a series of comparison experiments to exhibit the characteristics of Wheel-INS. Firstly, we illustrate the mounting angles' influence on the dead reckoning results. Then we show the DR results with different state dimensions in Wheel-INS and the covariance estimates of the inertial sensor error to demonstrate the analysis in Section III-A and explain why 21-state is used in Wheel-INS.

\subsubsection{Influence of Mounting Angles}

\begin{figure*}[htbp]
	\centering
	\subfigure[Estimated trajectories of Wheel-INS with/without mounting angle calibration against ground truth in Test 1.]{
		\includegraphics[width=8.5cm]{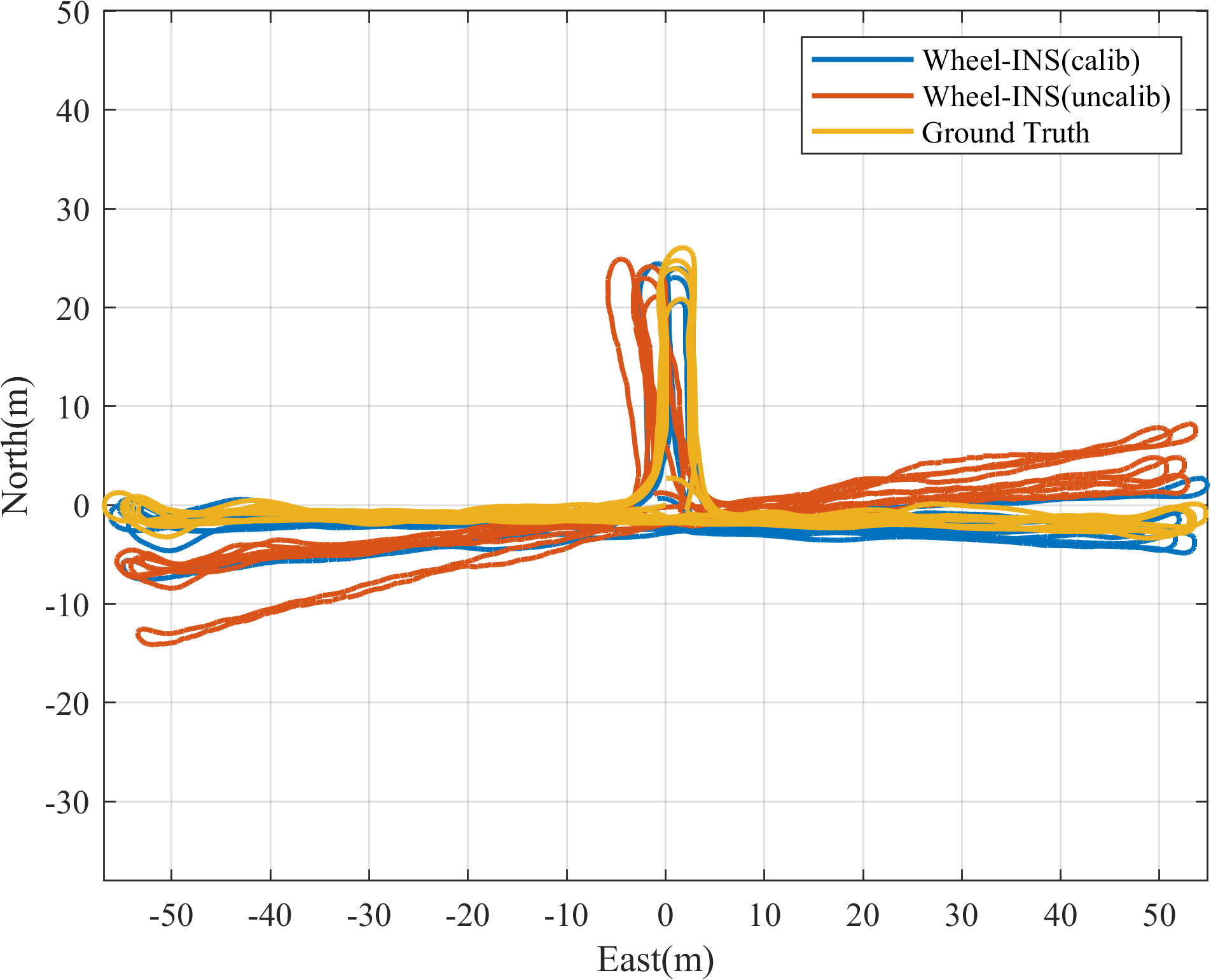}
	}
	\quad
	\subfigure[Positioning and heading errors of Wheel-INS with/without mounting angle calibration in Test 1.]{
		\includegraphics[width=8.5cm]{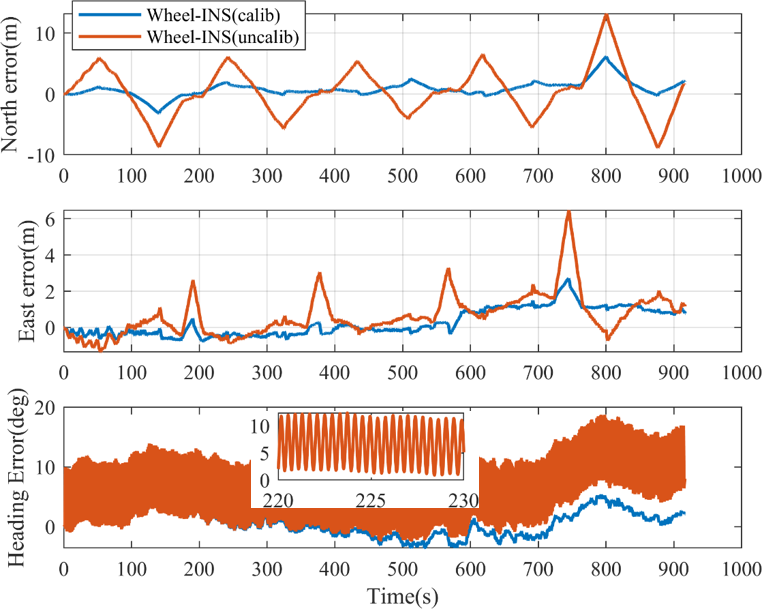}
	}
	\caption{Estimated trajectories and corresponding position and heading errors under different conditions in Test 1. ``calib'' indicates that the IMU mounting angles were calibrated and compensated beforehand, whereas ``noncalib'' indicates that they were not handled.}
	\label{fig7}
\end{figure*}

Fig. \ref{fig7} shows the calculated vehicle trajectory and the corresponding positioning and heading errors of Wheel-INS in Test 1 before and after compensating the mounting angles. The estimated values of the heading ($\delta\phi$) and pitch mounting angle ($\delta\theta$) in Test 1 were -4.5$^\circ$ and 2.5$^\circ$, respectively. It can be observed that the DR performance deteriorates when the mounting angles of Wheel-IMU are not compensated. As discussed in Section II-C, the reason is that the mounting angles weaken the rotation modulation effect and invalidate Eq. \ref{headingdifference} and Eq. \ref{vehicleeulerangle}, thereby introducing errors when fusing INS with the wheel velocity and NHC. From Eq. \ref{insvspeed} we can learn that given the same mounting angles, a higher vehicle velocity leads to a larger error of the INS-indicated wheel velocity in the \textit{v}-frame. Moreover, as shown in Fig. \ref{fig6}(b), if the mounting angles are not compensated in advance, a sine signal would be modulated onto the heading estimation. This is due to that the heading axis (\textit{x}-axis) of Wheel-IMU is not parallel to the rotation axis of the wheel, and the difference between them changes periodically with the rotation of the wheel.

\subsubsection{Comparison of Different State Dimension}

\begin{figure}[htbp]
	\centering
	\includegraphics[width=8.8cm]{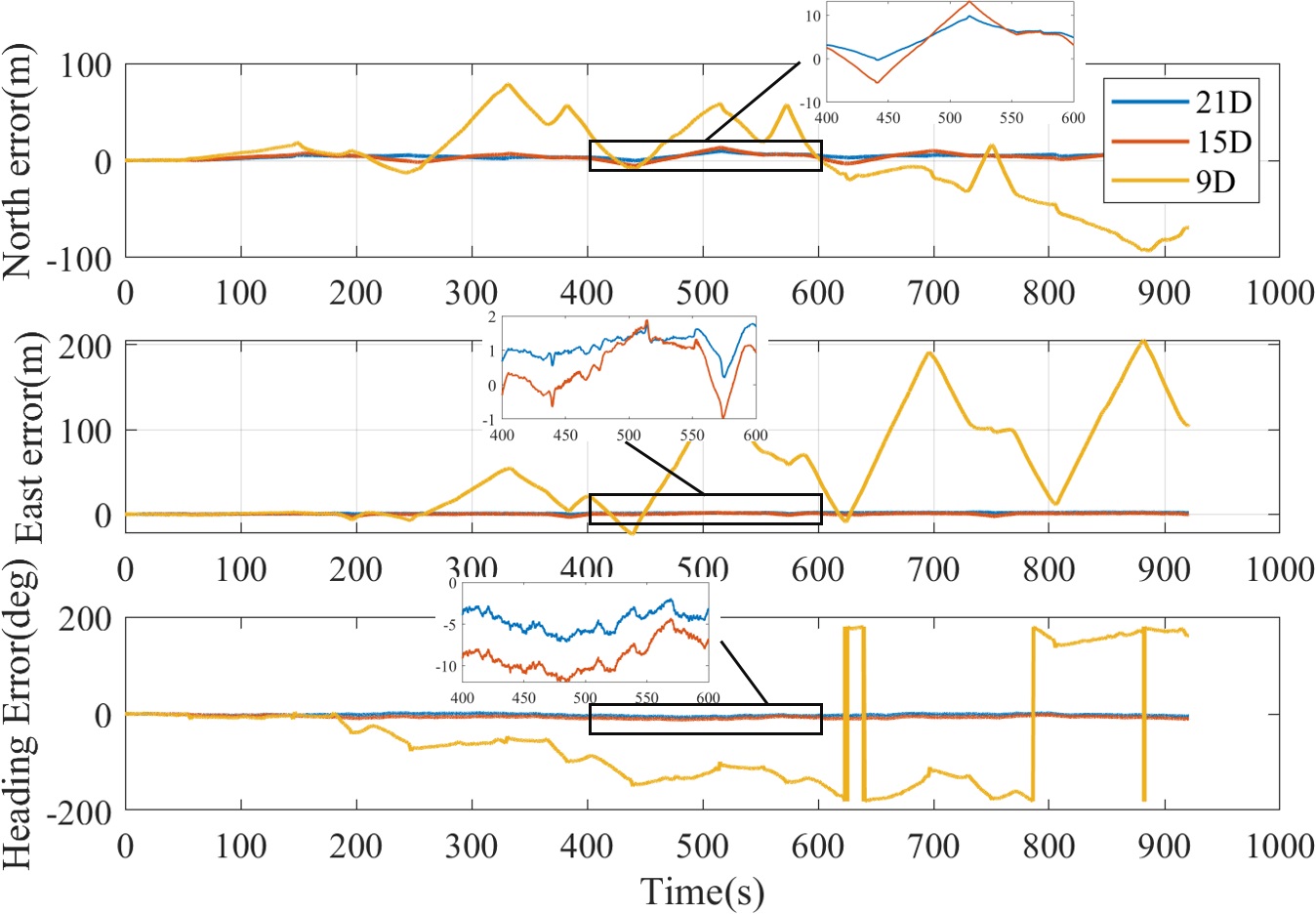}
	\caption{Comparison of the DR errors of Wheel-INS with different state dimensions in Test 1. ``21'', ``15D'' , and ``9D'' indicate the 21-, 15-, and 9-state, respectively.}
	\label{fig8}
\end{figure}

As discussed in Section III-A, we chose the 21-, 15-, and 9-dimensional state vectors in Wheel-INS respectively in Test 1 to compare their positioning performance. The 21-state is shown as Eq. \ref{statevector}. The 15-state does not include the scale factor error of the gyroscope and accelerometer, and the 9-state does not include all the IMU error terms. The corresponding navigation errors of the three experiments are shown in Fig. \ref{fig8}.

It can be observed from Fig. \ref{fig8} that the positioning error significantly increases when the dimension of the state vector is reduced to 9; the performance of the 15-state Wheel-INS is also worse than that of the 21-state Wheel-INS. Although the constant sensor bias of the IMU in the axis perpendicular to the rotation axis can be canceled to some extent, other error components can still cause large error if they are not estimated and compensated. For example, the cross-coupling errors have a more negative impact in Wheel-INS because the dynamic condition of Wheel-IMU is significantly higher than that of the IMU placed on the vehicle body in traditional ODO/INS.

\begin{figure}[htbp]
	\centering
	\includegraphics[width=8.8cm]{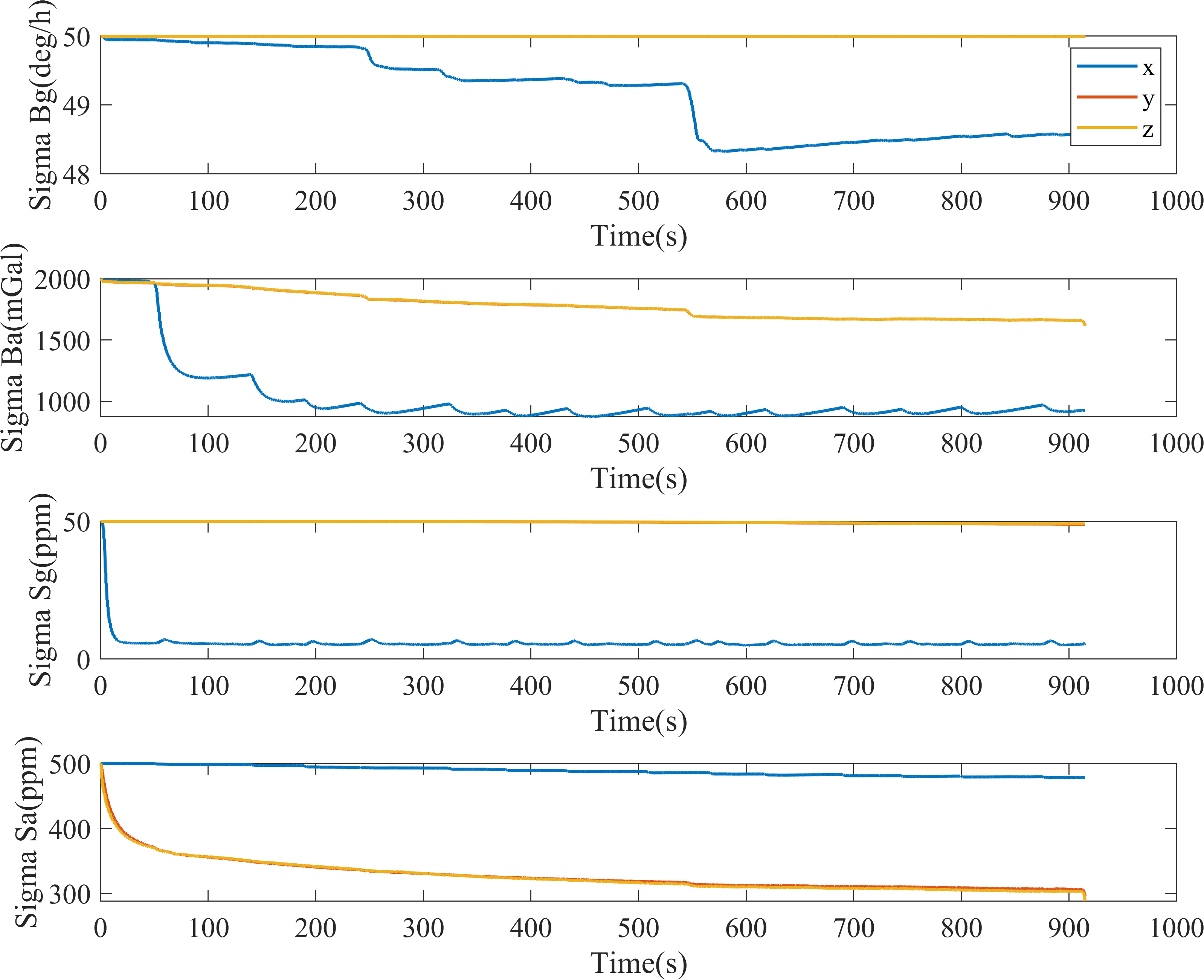}
	\caption{The STD of the estimates of the inertial sensor residual error in Wheel-INS (21-state) in Test 1.}
	\label{fig9}
\end{figure}

Fig. \ref{fig9} depicts the STD of the inertial sensor errors estimation in the 21-state Wheel-INS in Test 1. As discussed in Section III-A, the IMU errors in the \textit{y}- and \textit{z}-axes are mixed because they are parallel to the wheel plane and alternatively change their directions, making it difficult for the filter to distinguish them. This effect explains why the STD of the sensor error estimation in the \textit{y}- and \textit{z}-axes are almost coincident, as shown in all the subplots in Fig. \ref{fig9}. With regard to the scale factor error, because Wheel-IMU rotates around the \textit{x}-axis, a large rotation angular rate makes the gyroscopes scale factor error in the \textit{x}-axis observable. Therefore, it can soon converge in the filter system. In addition, as the accelerometers in the \textit{y}- and \textit{z}-axes perceive gravity alternatively and the vehicle is assumed to move on a horizontal plane, the  scale factor errors of the accelerometer in the \textit{y}- and \textit{z}-axes can be estimated by Wheel-INS.

\section{Conclusion}

A complete DR solution based on a wheel-mounted MEMS IMU is proposed in this study. The key objectives of this system is to: 1) only use one IMU to achieve the similar information fusion scheme as ODO/INS; 2) take advantages of the inherent rotation platform of the wheeled robot to spread the INS drift errors to all directions, so as to improve the ego-motion estimation accuracy. 

Field tests in different environments with different vehicles demonstrate that the maximum horizontal position drift of Wheel-INS is less than 1.8\% of the total traveled distance. The positioning and heading accuracy of Wheel-INS have respectively improved by 23\% and 15\% against ODO/INS. Moreover, benefit from the rotation modulation, Wheel-INS illustrates significant resilience to the gyroscope bias comparing with the conventional ODO/INS. Generally, a stabler wheel and axle structure and a relatively higher wheel speed would be beneficial for Wheel-INS.

The characteristics of Wheel-INS have also been analyzed and discussed. The misalignment errors of Wheel-IMU are defined and emphasized; they must be compensated in advance to obtain more robust and precise state estimates. Additionally, the observability of the inertial sensor errors are analyzed by their covariance propagated in the EKF. Because the residual sensor errors of the Wheel-IMU in the directions perpendicular to the rotation axis are coupled to each other, they cannot be distinguished and estimated by the EKF. Experimental results show that the 21-state EKF can achieve better performance. 

To promote Wheel-INS in practical long-term navigation applications of wheeled robots, one major issue need to be solved: power supply. Energy-harvesting techniques can be considered to achieve an ``install and forget'' solution. 

Future research directions include investigating approaches to use the vehicle attitude indicated by another IMU mounted on the vehicle body to extend Wheel-INS from 2D DR to 3D navigation. Furthermore, using two Wheel-IMUs mounted on left and right wheels is a promising design to obtain double information and take advantages of the spatial constraint between them to make the state estimation more robust.

\appendix

According to the error-state model in Eq. \ref{systemnodel} and Eq . \ref{sensorerrmodel}, the matrices $\textbf{F}(t)$ and $\textbf{G}(t)$ that appear in Eq. \ref{dynamicmodel} are given by

%\newcounter{TempEqCnt} % ??????TempEqCnt
%\setcounter{TempEqCnt}{\value{equation}} % ??????? ??TempEqCnt
%\setcounter{equation}{13} % ????????x?x???????????1.
%\begin{figure*}%?????????%ht???????
	\begin{equation}
	\textbf{F}(t) =\\
	\\
	\begin{bmatrix}
	\textbf{0}_{3} &\! \textbf{I}_3 &\! \textbf{0}_{3} &\! \textbf{0}_{3} &\! \textbf{0}_{3} &\! \textbf{0}_{3} &\! \textbf{0}_{3} \\
	\textbf{0}_{3} &\! \textbf{0}_{3} &\! \textbf{A} &\! \textbf{0}_{3} &\! \textbf{C}_b^n &\! \textbf{0}_{3} &\! \textbf{B} \\
	\textbf{0}_{3} &\! \textbf{0}_{3} &\! \textbf{0}_{3} &\! -\textbf{C}_b^n &\! \textbf{0}_{3} &\! \textbf{D} &\! \textbf{0}_{3}\\
	\textbf{0}_{3} &\! \textbf{0}_{3} &\! \textbf{0}_{3} &\! -\frac{1}{\mathrm{T}_{bg}}\textbf{I}_3 &\! \textbf{0}_{3} &\! \textbf{0}_{3} &\! \textbf{0}_{3}\\
	\textbf{0}_{3} &\! \textbf{0}_{3} &\! \textbf{0}_{3}  &\! \textbf{0}_{3} &\! -\frac{1}{\mathrm{T}_{ba}}\textbf{I}_3 &\! \textbf{0}_{3} &\! \textbf{0}_{3}\\
	\textbf{0}_{3} &\! \textbf{0}_{3} &\! \textbf{0}_{3} &\! \textbf{0}_{3} &\! \textbf{0}_{3} &\! -\frac{1}{\mathrm{T}_{sg}}\textbf{I}_3 &\! \textbf{0}_{3}\\
	\textbf{0}_{3} &\! \textbf{0}_{3} &\! \textbf{0}_{3} &\! \textbf{0}_{3} &\! \textbf{0}_{3} &\! \textbf{0}_{3} &\! -\frac{1}{\mathrm{T}_{sa}}\textbf{I}_3\\
	\end{bmatrix}
	\end{equation}
%\end{figure*}

\begin{equation}
\textbf{G}(t) =
\begin{bmatrix}
\textbf{0}_{3} &\! \textbf{0}_{3} &\! \textbf{0}_{3} &\! \textbf{0}_{3} &\! \textbf{0}_{3} &\! \textbf{0}_{3}\\
\textbf{C}_b^n &\! \textbf{0}_{3} &\! \textbf{0}_{3} &\! \textbf{0}_{3} &\! \textbf{0}_{3} &\! \textbf{0}_{3}\\
\textbf{0}_{3} &\! -\textbf{C}_b^n &\! \textbf{0}_{3} &\! \textbf{0}_{3} &\! \textbf{0}_{3} &\! \textbf{0}_{3}\\
\textbf{0}_{3} &\! \textbf{0}_{3} &\! \textbf{I}_3 &\! \textbf{0}_{3}  &\! \textbf{0}_{3} &\! \textbf{0}_{3}\\
\textbf{0}_{3} &\! \textbf{0}_{3} &\! \textbf{0}_{3}  &\! \textbf{I}_3 &\! \textbf{0}_{3} &\! \textbf{0}_{3}\\
\textbf{0}_{3} &\! \textbf{0}_{3} &\! \textbf{0}_{3} &\! \textbf{0}_{3} &\! \textbf{I}_3 &\! \textbf{0}_{3}\\
\textbf{0}_{3} &\! \textbf{0}_{3} &\! \textbf{0}_{3} &\! \textbf{0}_{3} &\! \textbf{0}_{3} &\! \textbf{I}_3\\
\end{bmatrix}
\end{equation}
where $\textbf{0}_{3}$ is the $3\times3$ zero matrix; $\textbf{I}_{3}$ is the $3\times3$ identity matrix; \textbf{A} is $(\textbf{C}_b^n\bm{f}^b)\times$; \textbf{B} is $\textbf{C}_b^{n}\mathrm{diag}(\bm{f}^b)$; \textbf{D} is $-\textbf{C}_b^n\mathrm{diag}(\bm{\omega}^b_{ib})$; $\mathrm{T}_{bg}$, $\mathrm{T}_{ba}$, $\mathrm{T}_{sg}$, $\mathrm{T}_{sa}$ are the correlation time of the Gauss-Markov processes corresponding to each inertial sensor error.

\section*{Acknowledgment}

The authors would like to thank Mr. Taiyu Li, Mr. Tao Liu and Mr. Hailiang Tang for their help in carrying out the experiments. 

\ifCLASSOPTIONcaptionsoff
  \newpage
\fi

{
\small
\bibliographystyle{IEEEtran}
\bibliography{IEEEabrv, ReferenceWheel-INS}
}
%\vspace{-200 mm}

\begin{IEEEbiography}[{\includegraphics[width=1in,height=1.25in,clip,keepaspectratio]{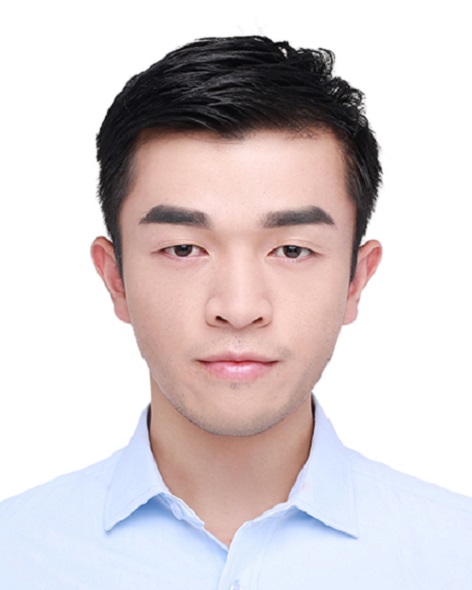}}]{Yibin Wu} received the B.Eng. degree (with honors) in Navigation Engineering and the M. Eng. degree from Wuhan University, Wuhan, China, in 2017 and 2020, respectively. His research interests focus on inertial navigation, multi-sensor fusion, and mobile robot state estimation.
\end{IEEEbiography}
%\vspace{-100 mm}
\begin{IEEEbiography}[{\includegraphics[width=1in,height=1.25in,clip,keepaspectratio]{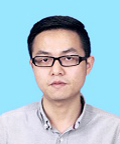}}]{Jian Kuang} received the B.Eng. degree and Ph.D. degree in Geodesy and Survey Engineering from Wuhan University, Wuhan, China, in 2013 and 2019, respectively. He is currently a Postdoctoral Fellow with the GNSS Research Center in Wuhan University, Wuhan, China. His research interests focus on inertial navigation, pedestrian navigation, and indoor positioning.
\end{IEEEbiography}
%\vspace{-100 mm}
\begin{IEEEbiography}[{\includegraphics[width=1in,height=1.25in,clip,keepaspectratio]{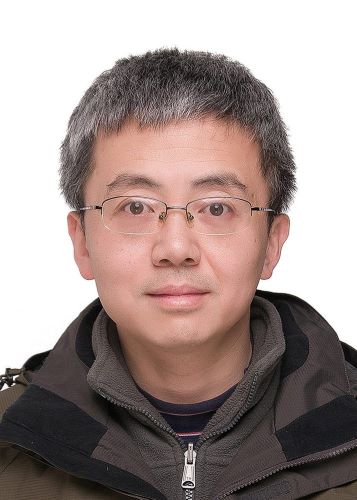}}]{Xiaoji Niu} received the B.Eng. degree (with honors) in Mechanical Engineering and the Ph.D. from Tsinghua University, Beijing, China, in 1997 and 2002, respectively. From 2003 to 2007, he was a Post-Doctoral Fellow with the Department of Geomatics Engineering, University of Calgary. From 2007 to 2009, he was a senior scientist with SiRF Technology, Inc. He is currently a Professor of the GNSS Research Center and the Artificial Intelligence Institute of Wuhan University, Wuhan, China. His research interests focus on INS, GNSS/INS integration for land vehicle navigation and pedestrian navigation.
\end{IEEEbiography}


\end{document}